%% file: icmlW_2025.tex
\documentclass{article}

\usepackage{hyperref}       
\usepackage{url}            
\usepackage{booktabs}       
\usepackage{amsfonts}       
\usepackage{nicefrac}       
\usepackage{microtype}      
\usepackage{xcolor}         
\usepackage{graphicx}
\usepackage{svg}       
\usepackage[most]{tcolorbox}
\usepackage[most]{tcolorbox}
\usepackage{array}
\usepackage{tabularx}
\usepackage{subcaption}
\usepackage{colortbl}
\usepackage{multirow}
\usepackage{booktabs}
\usepackage{placeins}
\usepackage{tabularx}
\usepackage{fontawesome5}
\usepackage{float}

\usepackage[accepted]{icml2025}

\usepackage{amsmath}
\usepackage{amssymb}
\usepackage{mathtools}
\usepackage{amsthm}
\usepackage{tcolorbox}
\usepackage{fontawesome5}

\usepackage[capitalize,noabbrev]{cleveref}

\theoremstyle{plain}

\theoremstyle{definition}

\theoremstyle{remark}

\usepackage[textsize=tiny]{todonotes}
\tcbuselibrary{listings}
\graphicspath{{./Figures/}}

\icmltitlerunning{Supernova Event Dataset: Interpreting Large Language Model's Personality through Critical Event Analysis}

\newtcolorbox{promptbox}{
  enhanced,
  colback=gray!5!white,
  colframe=gray!75!black,
  fonttitle=\bfseries,
  title=Prompt,
  arc=0mm,
  boxrule=0.5pt
}

\begin{document}

\twocolumn[
\icmltitle{Supernova Event Dataset: Interpreting Large Language Models' Personality through Critical Event Analysis}




\icmlsetsymbol{equal}{*}

\begin{icmlauthorlist}
\icmlauthor{Pranav Agarwal}{equal,zzz}
\icmlauthor{Ioana Ciucă}{equal,yyy}
\end{icmlauthorlist}

\icmlaffiliation{yyy}{Stanford University}
\icmlaffiliation{zzz}{Mila, Quebec AI Institute}

\icmlcorrespondingauthor{Pranav Agarwal}{pranav.agarwal.2109@gmail.com}
\icmlcorrespondingauthor{Ioana Ciucă}{iciuca@stanford.edu}


\icmlkeywords{Machine Learning, ICML}

\begin{center}
\tcbset{
    on line,
    boxsep=1pt, left=8pt, right=8pt,
    colback=white,
    colframe=white,
    boxrule=0pt,
}
\tcbox{%
    \href{http://supernova-event.ai/}{\faGlobe~\textbf{Website}}
    \qquad
    \href{https://github.com/pranavAL/Supernova-Event-Dataset}{\faGithub~\textbf{Code}}
    \qquad
    \href{https://huggingface.co/datasets/SupernovaEvent/SupernovaEventDataset}{%
        \raisebox{-0.7ex}{\includegraphics[height=2.8ex]{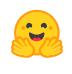}}~\textbf{Dataset}}
}
\end{center}

\vskip 0.1in
]

\printAffiliationsAndNotice{\icmlEqualContribution}

\input{0_abstract}

\input{1_intro}

\input{2_Related_work}

\input{3_dataset}

\input{4_methodology}

\input{5_evaluation}

\input{6_Results}

\input{7_ablation}

\input{8_conclusions}

\input{bibliography}


\input{10_appendix}

\end{document}

%% file: 0_abstract.tex
\begin{abstract}
Large Language Models (LLMs) are increasingly integrated into everyday applications. As their influence grows, understanding their decision-making and underlying personality becomes essential. In this work, we interpret model personality using our proposed \textit{Supernova Event} Dataset, a novel dataset with diverse articles spanning biographies, historical events, news, and scientific discoveries. We use this dataset to benchmark LLMs on extracting and ranking key events from text, a subjective and complex challenge that requires reasoning over long-range context and modeling causal chains. We evaluate small models like Phi-4, Orca 2, and Qwen 2.5, and large, stronger models such as Claude 3.7, Gemini 2.5, and Open AI o3, and propose a framework where another LLM acts as a judge to infer each model’s personality based on its selection and classification of events. Our analysis shows distinct personality traits: for instance, Orca 2 demonstrates emotional reasoning focusing on interpersonal dynamics, while Qwen 2.5 displays a more strategic, analytical style. When analyzing scientific discovery events, Claude Sonnet 3.7 emphasizes conceptual framing, Gemini 2.5 Pro prioritizes empirical validation, and o3 favors step-by-step causal reasoning. This analysis improves model interpretability, making them user-friendly for a wide range of diverse applications.
  
\end{abstract}

%% file: 1_intro.tex
\section{Introduction}
\label{sec:introduction}

\begin{figure*}[t]
    \centering
    \includegraphics[width=\textwidth]{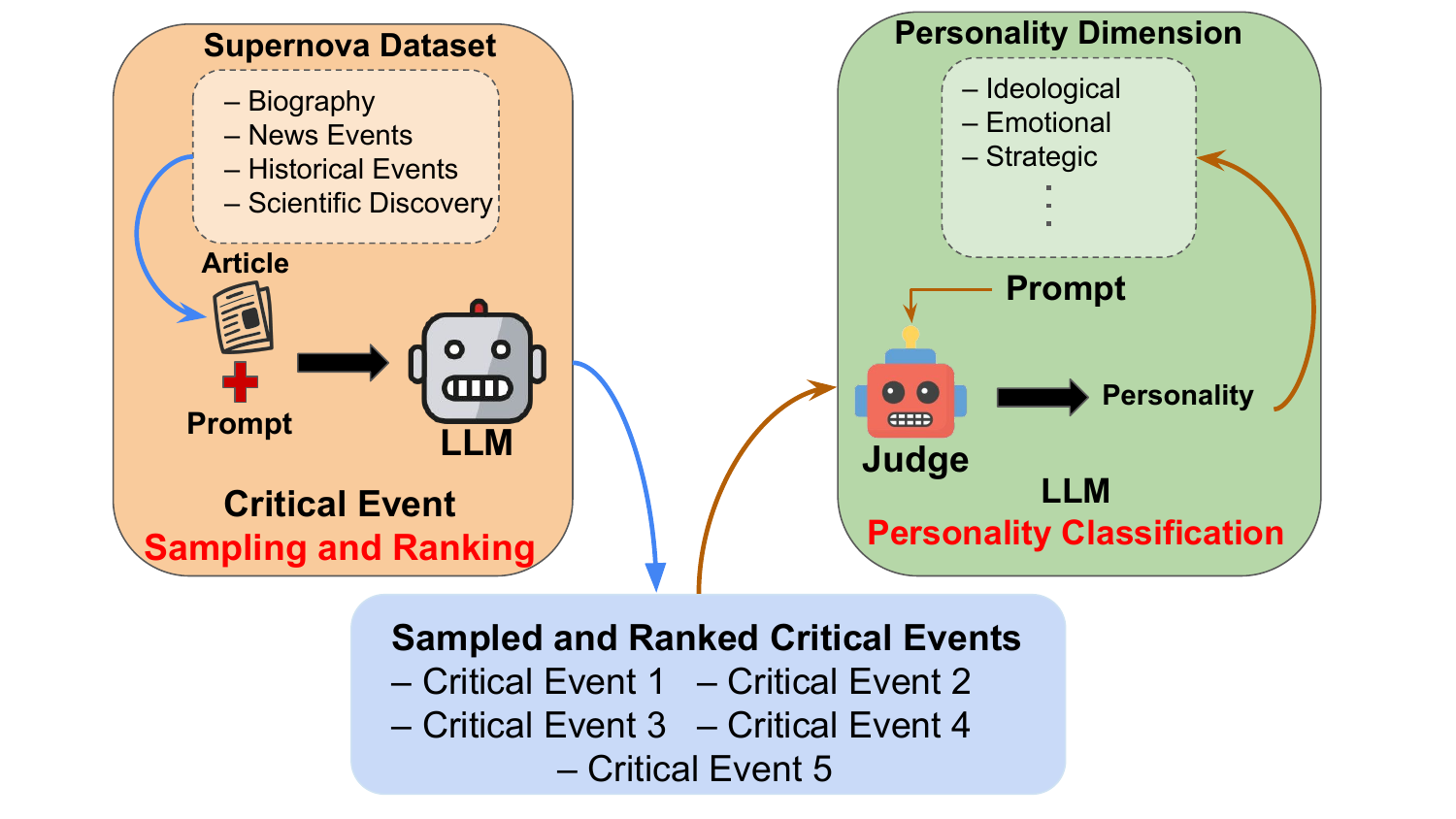}
    \caption{Overview of our LLM personality analysis framework. The framework utilizes our Supernova event dataset, a collection of Wikipedia biographies, major news, historical events, and scientific discoveries from Google Deep Research. The target LLM receives an article from this corpus (via RAG) along with a prompt, then samples and ranks the five most critical events in order of importance. A judge LLM analyzes these selections and rankings to determine the target LLM's personality, revealing its human values and decision-making patterns. (\textit{We use ``personality" to describe consistent behavioral patterns, not to imply consciousness or emotion})}
    \label{fig:intro-fig}
\end{figure*}

As Large Language Models (LLMs) become integrated in high-stakes domains such as healthcare, law, finance, and education, evaluating their capabilities beyond factual accuracy becomes crucial. Most current LLM benchmarks focus on objective tasks with verifiable ground truths~\citep{ivanov2024ai}, but these are becoming insufficient to capture subjective judgment, interpretation, and alignment with human values~\citep{aoyagui2025matter}.

A subjective task such as identifying and ranking critical events in a detailed timeline of an event (for example, a biography, historical and news events, or scientific discovery) is challenging~\citep{stranisci2022guidelines, plum2019large}. Selecting the most important critical event requires \textit{reasoning} across different levels of abstraction and understanding long-term dependencies~\citep{kourani2022mining}. This is challenging due to human memory limitations and the non-linear interactions between events, which involve subtle causal relationships~\citep{gianicolo2020methods}. For the same event, humans can sample different critical events based on their values and experiences, often leading to disagreement. 

This task resembles the concept of salience detection in cognitive science, where information stands out due to both intrinsic properties and relevance to perceived goals~\citep{liu2018automatic}. In natural language processing (NLP) research, event salience has been explored through filtering non-salient events, leveraging contextual information, and examining how an event's removal affects narrative coherence~\citep{zhang2021salience, otake2020modeling}.

For LLMs, identifying the most critical events is especially challenging, as they lack the inherent understanding of events that humans possess~\citep{belisle2024we, ding2024language}. The ranking of these sampled events adds another layer of difficulty, as it requires complex reasoning, understanding causal relationships that are often not explicitly stated, and interpreting many subtle factors~\citep{su2025enhancing, cai2025dr, li2025survey}. Each LLM exhibits a unique reasoning style shaped by its pre- and post-training strategies~\citep{besta2025reasoning, kumar2025llm}. Our dataset and task formulation make it possible to uncover human values and, in turn, personality traits reflected in each model’s analysis of critical events.

Although previous work has shown that LLMs can simulate personality traits when explicitly prompted~\citep{sorokovikova2024llms}, our work demonstrates that even without role-playing prompts, LLMs exhibit consistent style in their decision-making that reveals unique personality traits when handling complex subjective tasks such as critical event analysis.

In this work, we introduce the Supernova Event Dataset, a collection of primarily Wikipedia articles covering biographies, news, historical events, and scientific discoveries. These categories were chosen for their clear timelines and multiple key events. Using this dataset, we define a task that involves sampling and ranking critical events from the given article. Another LLM is used as a judge to assess the selected and ranked events to evaluate the personality of various Large Reasoning Models (LRMs) and Large Language Models (LLMs). We also include an ablation study (Sec:~\ref{sec:ablation}) using a movie script dataset to examine the behavior of the model in a different domain.

The main contributions of this work are:
\begin{itemize}
    \item \textbf{Supernova Event Dataset:} We introduce a new dataset consisting of Wikipedia articles on diverse topics, including biographies, news, historical events, and scientific discoveries from Google Deep Research. In addition to enabling personality interpretation, this dataset can help future research evaluate how well models handle long context, multi-dimensional causal chains, and counterfactual reasoning.
    
    \item \textbf{New Task for Personality Benchmarking:} A novel task of critical event sampling and ranking is introduced to benchmark the personality traits of LLMs. This task is prompt-agnostic, which makes it effective in revealing consistent model behavior.

    \item \textbf{Personality Evaluation Framework:} We propose a framework to evaluate the personality of a model by using an additional LLM as a judge. The judge assesses the sampled and ranked critical events produced by the candidate model, helping to avoid human biases and cognitive overload.

\end{itemize}

We acknowledge that using ``personality" to describe LLM behavior is metaphorical as LLMs lack consciousness and emotions. However, following recent work~\citep{he2025investigating, wang2025evaluating}, we find personality a useful framework for characterizing consistent behavioral patterns. 

Our work is the first to extract the precise personality of LLMs in a more realistic setup. Our analysis shows that LLM decision-making can be made more interpretable by designing better evaluation methods and tasks, which is important for their safe deployment.

%% file: 2_Related_work.tex
\section{Related Work}
\label{sec:related}

\begin{table*}[!t]
\centering
\small
\caption{Comparison of models' ranking of critical events in Subrahmanyan Chandrasekhar's life}
\label{tab:model_bio}
\begin{tabular}{>{\raggedright\arraybackslash}p{0.31\textwidth}>{\raggedright\arraybackslash}p{0.31\textwidth}>{\raggedright\arraybackslash}p{0.31\textwidth}}
\toprule
\textbf{phi4} & \textbf{orca2} & \textbf{qwen 2.5} \\
\midrule
1. Nobel Prize (1983) & 1. Chandrasekhar limit discovery & 1. Chandrasekhar limit discovery \\
2. Dispute with Eddington (1930s) & 2. Trinity College Fellowship & 2. Nobel Prize (1983) \\
3. U. Chicago appointment (1936) & 3. Nobel Prize (1983) & 3. Dispute with Eddington \\
4. WWII ballistics research & 4. U. Chicago professorship & 4. WWII ballistics work \\
5. Early education/family influence & 5. Hertzsprung-Russell diagram & 5. Philosophy of Systematization \\
\bottomrule
\end{tabular}
\vspace{1pt}
\caption*{\footnotesize \textbf{Model perspective:} Models show different priorities in ranking significant events, varying in temporal ordering and professional vs. personal achievement emphasis.}
\end{table*}

LLMs such as GPT-3.5 and GPT-4 can simulate personality traits when prompted to assume specific roles, and these traits are often recognizable to human evaluators based on the generated content~\citep{jiang2023personallm}. This ability has been shown in tasks such as answering questionnaires~\citep{wang2025evaluating}. Similarly, LLMs tend to follow certain moral competence~\citep{khamassi2024strong} to avoid harmful content and reflect the social biases present in the training data~\citep{deng2024deconstructing}. However, our work shows that even without role-playing prompts, LLMs tend to follow consistent reasoning patterns and reflect certain human values, revealing a unique personality of their own when handling complex, subjective tasks like critical event analysis. 

\citet{heston2025large} and \citet{bodrovza2024personality} adapted psychological tools such as the Big Five Inventory and Schwartz’s values to measure behavioral patterns or ``traits" in LLM. Their results show that different LLMs can exhibit distinct profiles. Recent work has explored LLM personality through various lenses. \citet{he2025investigating} showed LLMs can adopt Big Five personality traits when prompted, while~\citet{wang2025evaluating} demonstrated these traits affect decision-making patterns. However, these approaches typically use explicit personality framing or psychometric tests. Our work examines whether consistent personality-like patterns emerge naturally in complex tasks without such prompting.

Event extraction is important to improve the accuracy and reliability of LLMs when handling complex, long-form texts. Recent benchmarks on long-context reasoning~\citep{ling2025longreason, kuratov2024babilong} show that models often struggle with tasks that require integrating information across extended passages, highlighting the challenging nature of our task setup.

Recent work has explored the application of LLMs for the extraction of events from long-form text. \citet{liu2024document} introduced Definition-driven Document-level Event Extraction (DDEE), which enhances prompt design and incorporates structured heuristics. This approach addresses challenges such as prompt sensitivity and the long-tail distribution of event types. 

\citet{zhang2024ultra} proposed ULTRA, a hierarchical framework that efficiently extracts event arguments from entire documents. ULTRA mitigates positional bias by processing text in chunks. \citet{gao2024eventrl} proposes EventRL, which applies reinforcement learning to train LLMs for better event extraction. By introducing specific reward functions, EventRL enhances the model's adherence to instructions and reduces hallucinations, particularly in handling novel event types.

Although most existing work focuses on improving the accuracy of event extraction from long-form text, our work introduces a new, subjective task: identifying and ranking the most critical event. This task shifts the focus from extraction accuracy to how LLMs prioritize events based on their importance, offering a deeper insight into the personality traits and human values reflected by each model.

%% file: 3_dataset.tex
\section{Supernova Event Dataset}
\label{sec:dataset}

The Supernova Event dataset includes Wikipedia articles on biographies, historical and major news events, and scientific discoveries created using Gemini 2.5 Pro Deep Research. Articles are chosen based on criteria such as word count or the frequency of edits, prioritizing those with the highest number of edits as an indicator of importance (Table~\ref{tab:corpus_overview}). We collect only the text content, excluding all other modalities.

\subsection{Biography Dataset Collection}

Our biographies collection pipeline targets entries with standardized infobox templates, including person, scientist, writer, actor, politician, and sports personnel. We employ strict content filtering as followed in the work by~\citet{shao2024assisting}, which requires a minimum of 3,000 words to ensure a comprehensive coverage of the person's life and achievements. The crawler's template-based categorization approach eliminates disambiguation pages and other nonbiographical content, while our page view threshold ($>$50,000 views) ensures we capture only notable figures with significant public interest. Each biography is parsed using \texttt{mwparserfromhell} to remove the wiki markup and extract clean, readable text before being saved.

\subsection{Historical and World News Events Dataset Collection}
We compile two additional categories consisting of \emph{Major World News Events (2000 -- present)} and \emph{Global Historical Turning-Points (1000 BCE -- 2000 CE)} from Wikipedia. Each crawler walks a curated set of high-level categories (e.g., \emph{21st-century conflicts}, \emph{Battles}, \emph{Disasters}) to a depth of 1, then filters candidates using: (i) basic heuristics that reject year-only, list, disambiguation, and slogan pages, (ii) filters on article length (word count $\geq$ 500), ORES quality class ($\geq$ B), and cumulative page-views ($\geq$ 5000), and (iii) explicit year extraction confined to the temporal window.

Articles that pass these filters are further processed for semantic validation, where a local LLM (LLaMa-3-8B served through \texttt{ollama}) with a 1500 token context is queried to ensure if the page is \emph{primarily about a discrete major event}. A given article passes if the LLM classifies it as an event with confidence greater than 0.9. The seed articles are automatically accepted. Finally, human verification is performed, and 200 articles are curated for each category.

\subsection{Scientific Discovery Dataset Collection}
The endpoints \emph{Physics}, \emph{Chemistry}, and \emph{Physiology or Medicine} (category codes \textit{phy}, \textit{che}, and \textit{med}) are queried using the Nobel Prize REST API~v2.1. For each prize record, the award year, the English category name, the English motivation text, and the list of laureate names are extracted. The query results in 384 prizes between 1901 – 2024, organized in a JSON object with a four‐field schema (\textit{year}, \textit{category}, \textit{discovery}, \textit{laureates}).

We prompt Google Gemini 2.5 Pro with Deep Research to transform Nobel entries into appropriate narrative instances for long-context reasoning. Each prompt contains the raw Nobel metadata, and Gemini returns a fully formed encyclopedic article that covers historical context, methodology, publication trail, significance, and legacy. We do not apply any post-processing and verify for hallucination before saving our articles as text files. We generate 25 expanded scientific discovery articles.

\begin{table*}[!htbp]                       
  \small
  \setlength{\tabcolsep}{4pt}           
  \renewcommand\arraystretch{1.1}       
  \caption{The Supernova Event dataset}
  \label{tab:corpus_overview}
  \centering

  \begin{tabularx}{\textwidth}{@{}
        l                                   
        >{\raggedright\arraybackslash}X     
        c                                   
        c                                   
        >{\raggedright\arraybackslash}X     
        r                                   
        r@{}}                               
    \toprule
    \textbf{Category} & \textbf{Source} & \textbf{Min\;Words} &
    \textbf{Min\;Page-views} & \textbf{Additional Filters} &
    \textbf{\# Articles} & \textbf{\# Processed} \\ \midrule

    Biographies &
      English Wikipedia &
      3000 & 50000  &
      n/a &
      192 & 150 \\

    Historical Events &
      English Wikipedia &
      500 & 5000 &
      ORES `B' \& LLM \& Year $<$ 2000 &
      200 & 150 \\

    Major News Events &
      English Wikipedia &
      500 & 5000 &
      ORES `B' \& LLM \& Year $>$ 2000 &
      200 & 150 \\

    Scientific Discoveries &
      Gemini Deep Research &
      n/a & n/a & 
      n/a & 25 & 25 \\ \bottomrule
  \end{tabularx}
\end{table*}


%% file: 4_methodology.tex
\section{Methodology}
\label{sec:method}

To systematically identify and rank critical events within a given document, we use Retrieval-Augmented Generation (RAG). The corpus (comprising Wikipedia articles) is processed using a chunking mechanism that segments lengthy documents into smaller, semantically coherent units (1000 tokens with 100 tokens overlap). Each segment is then transformed into a high-dimensional vector representation using the \textit{nomic-embed-text-v1} embedding model~\citep{nussbaum2024nomic}, creating a searchable semantic space.

The embedded document chunks are indexed in a \textit{FAISS} vector database~\citep{johnson2019billion}, which facilitates efficient search across the entire corpus. This approach provides a significant computational advantage over traditional search methods when working with large-scale document collections. For each document, a context-aware retrieval system filters and ranks the most relevant text chunks based on semantic similarity to the query. Our approach utilizes a two-stage prompting strategy, where the first prompt is specifically designed to enhance the retrieval capabilities of the system as elaborated in Section~\ref{sec:prompts} (Box 1).

The initial prompt enables the MultiQueryRetriever to reformulate the original query into multiple search queries, thereby increasing the likelihood of retrieving semantically relevant text chunks containing critical event information. By explicitly instructing the model to consider factors that define critical events (such as turning points and cascading effects), the retriever is optimized to locate passages containing significant milestones rather than topic-related content.

Once the relevant document chunks are retrieved, a second prompt guides the large language model through a structured analytical process, further elaborated in Section~\ref{sec:prompts} (Box 2).

For the scientific discovery corpus, each article is used for analyzing the strong LRMs, o3, Gemini 2.5 Pro, and Claude Sonnet 3.7 Thinking. Each model is prompted to identify and rank the five turning points that most decisively altered the trajectory toward the given discovery or its recognition, using explicit counterfactual tests (“Would the narrative have unfolded differently?”) as selection criteria. The model returns an ordered list with summaries of one sentence and concludes with a reflective label that represents the guiding principle behind its classification. We also keep track of the chains of reasoning that human experts can examine in later evaluation stages.

\begin{table*}[!htbp]
\centering
\small
\caption{Comparison of models' ranking of critical events in the 2008 Financial Crisis}
\label{tab:financial_crisis_comparison}
\begin{tabular}{>{\raggedright\arraybackslash}p{0.31\textwidth}>{\raggedright\arraybackslash}p{0.31\textwidth}>{\raggedright\arraybackslash}p{0.31\textwidth}}
\toprule
\textbf{phi4} & \textbf{orca2:13b} & \textbf{qwen2.5:14b} \\
\midrule
1. Lehman Brothers Bankruptcy & 1. Collapse of Lehman Brothers & 1. Subprime Mortgage Crisis \\
2. Failure of AIG and Government Bailout & 2. Seizure of Fannie Mae and Freddie Mac & 2. Collapsing Housing Market \\
3. Collapse of Bear Stearns & 3. Passage of TARP & 3. Deregulation of Financial Markets \\
4. Passage of TARP & 4. Quantitative easing by Federal Reserve & 4. AIG Collapse and Bailout \\
5. Subprime Mortgage Crisis Peak & 5. European debt crisis & 5. Global Impact \\
\bottomrule
\end{tabular}
\vspace{2pt}
\caption*{\footnotesize \textbf{Model approaches:} Phi4 and Orca2 both rank the Lehman Brothers collapse as most critical, focusing on immediate catalysts, while Qwen2.5 emphasizes underlying causes by ranking the Subprime Mortgage Crisis first. Phi4 and Orca2 prioritize specific institutional failures, while Qwen2.5 takes a more systemic, macro-level approach to the crisis narrative.}
\end{table*}

%% file: 5_evaluation.tex
\section{Evaluation}
\label{sec:eval}

To evaluate the sampled critical point and its ranking thereafter benchmarking the personality of different language models, we use an external LLM as a judge. We use Qwen-2.5 (14B) as a judge. The evaluation compared three alternative models: Phi-4 (LRM), Orca-2 (LRM), and Qwen-2.5 (LLM).

\subsection{Benchmarking Model Personality}

To evaluate or label the personality of an LLM in solving the given task, we use a meta-analysis technique where one LLM (specifically \textit{qwen 2.5:14b} as the analysis model) evaluates the personality of other LLMs (phi4, orca2:13b, and qwen 2.5:14b) based on their outputs (see Fig.~\ref{fig:intro-fig}). Our indirect approach to personality assessment is motivated by recent findings that LLMs' self-explanations often misrepresent their actual reasoning processes~\citep{lindsey2025biology}. Rather than asking models to self-report personality traits, we observe their behavior in complex tasks. 

Each target model analyzes the given text and identifies and ranks critical events. The analysis model is then prompted to examine the target model's output using a structured evaluation framework. The analysis model synthesizes the output into a concise ``personality type". This use of an external LLM for analysing avoids the requirement of heuristics, which are hard to generalize, providing a scalable method. 

For using this external model as a judge to evaluate the other models, we use the prompt as specified in Section~\ref{sec:prompts} (Box 3) along with the full response of the target language model being evaluated as input for the LLM judge (Fig.~\ref{fig:intro-fig}). While this provides consistency and scalability, it introduces potential biases and lacks human validation. Future work should incorporate human evaluation and explore multi-judge agreement to validate these findings. Additionally, our personality categories are empirically derived rather than grounded in established psychological frameworks. We view this as exploratory research that opens new avenues for understanding LLM behavior.

\subsection{Personality Trait Identification}
To identify personality traits from LLM responses, we use a sentence transformer model to generate semantic embeddings of the traits, enabling both similarity measurement and dimensionality reduction. 

Specifically, we employ the \textit{all-MiniLM-L6-v2} model from the sentence-transformers library to encode each identified personality trait into a dense vector representation. For each LLM, we compute an aggregate embedding by combining the embeddings of its associated traits, weighted by their frequency. Cosine similarity between these aggregate embeddings allows us to quantify personality similarity across models. Finally, we apply Principal Component Analysis (PCA) to reduce the embedding space to two dimensions for visualization purposes.

%% file: 6_Results.tex
\section{Results}
\label{sec:results}

\begin{table*}[htb]
\centering
\small
\caption{Comparison of models' ranking of foundational discoveries enabling machine learning with ANNs}
\label{tab:ml_discoveries_comparison}
\begin{tabular}{>{\raggedright\arraybackslash}p{0.31\textwidth}>{\raggedright\arraybackslash}p{0.31\textwidth}>{\raggedright\arraybackslash}p{0.31\textwidth}}
\toprule
\textbf{o3} & \textbf{Gemini 2.5 Pro} & \textbf{Claude Sonnet 3.7 Thinking} \\
\midrule
1. Hopfield's 1982 energy-based network paper & 1. Backpropagation (1986 Nature Paper) & 1. Backpropagation Paper in Nature (1986) \\
2. 1986 Nature backpropagation breakthrough & 2. Hopfield's Application to Associative Memory (1982) & 2. Hopfield's Network Model in PNAS (1982) \\
3. 1985 Boltzmann-machine formulation & 3. Boltzmann Machine (1985) & 3. Minsky \& Papert's Perceptrons Book (1969) \\
4. 1969 Perceptrons critique and AI Winter & 4. Minsky \& Papert's Perceptrons Critique (1969) & 4. Boltzmann Machine Paper (1985) \\
5. 2024 Nobel Prize announcement & 5. Explicit Use of Physics Principles & 5. Rosenblatt's Perceptron (late 1950s) \\
\bottomrule
\end{tabular}
\vspace{2pt}
\caption*{\footnotesize \textbf{Model perspectives:} o3 uniquely includes the 2024 Nobel Prize while being the only model to rank Hopfield's work first. Both Gemini and Claude rank the 1986 backpropagation paper as most significant. Claude uniquely includes Rosenblatt's original perceptron work from the 1950s, giving historical context absent in other models' rankings.}
\end{table*}

 We observe consistent differences in how models approach event selection, which we characterize using descriptive labels.

\subsection{Personality Category Distribution}

\begin{figure*}[htbp]
\centering
\begin{subfigure}{.5\textwidth}
  \centering
  \includegraphics[width=\linewidth]{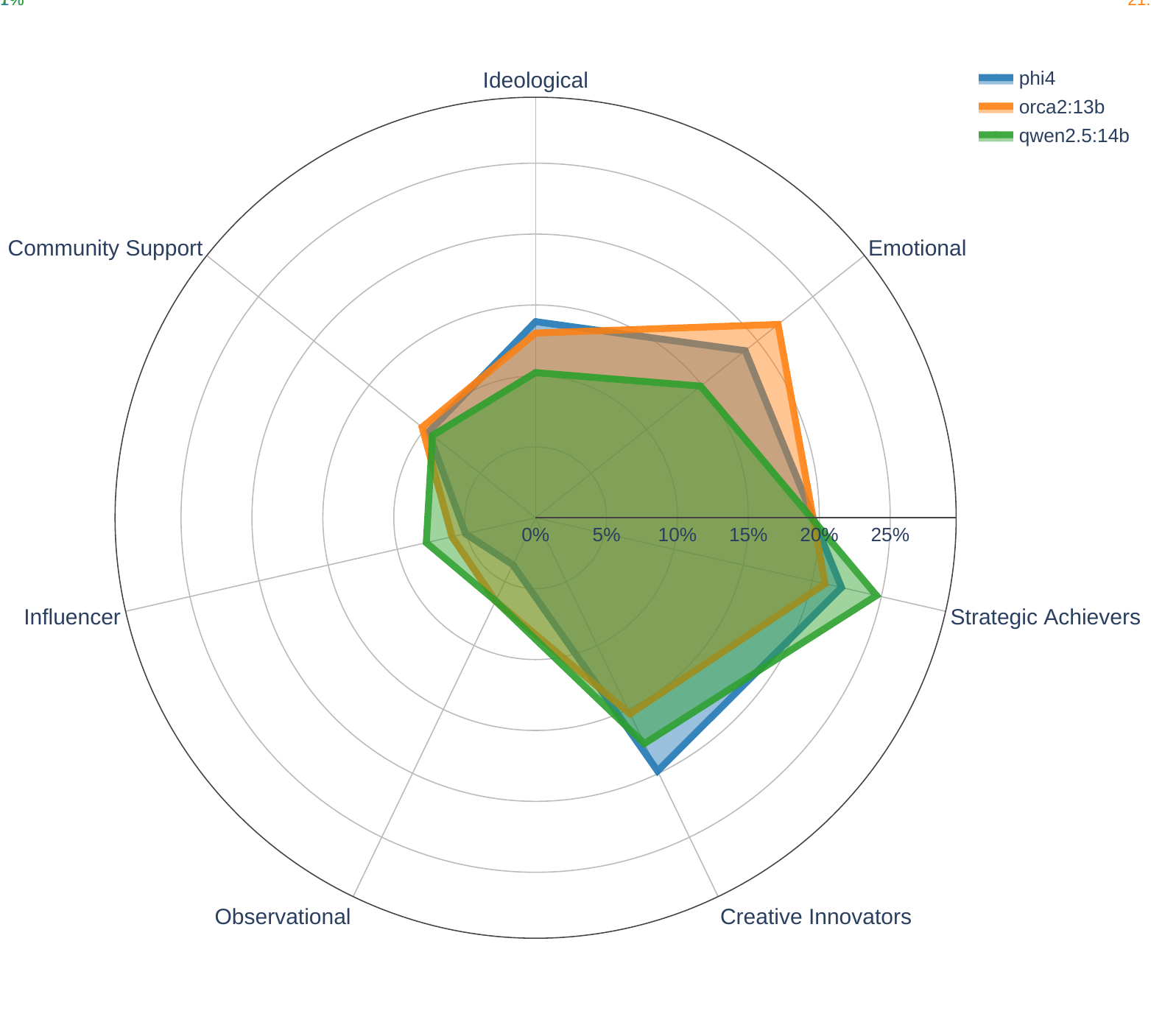}
  \caption{Personality category distribution}
  \label{fig:radar}
\end{subfigure}%
\begin{subfigure}{.5\textwidth}
  \centering
  \includegraphics[width=\linewidth]{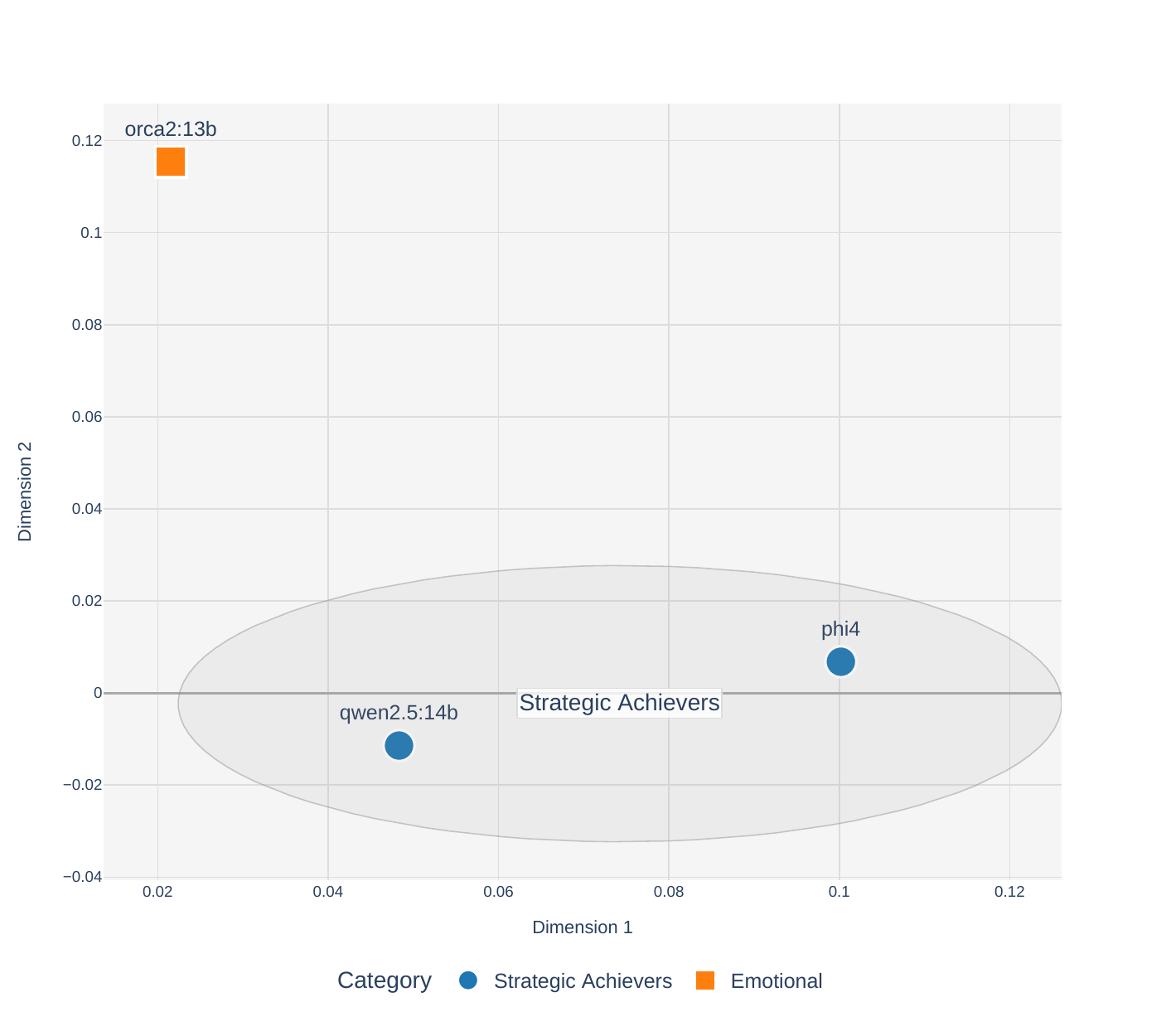}
  \caption{Model semantic space}
  \label{fig:semantic}
\end{subfigure}
\caption{Analysis of LLM personality profiles. (a) Shows the distribution of personality categories for each model, with higher values indicating stronger presence. (b) Positions models in 2D semantic space based on their personality traits.}
\label{fig:all-plots}
\end{figure*}

Figure~\ref{fig:radar} presents each model’s distribution across seven personality categories, revealing distinct profiles. Phi4 stands out for its strong ``Strategic Achievers” and ``Creative Innovators” traits, with moderate emotional presence and lower scores on Ideological, Observational, and Influencer dimensions. Orca2:13b exhibits the highest ``Emotional” score of the three and a modest uptick in ``Community Support,” alongside solid strategic ability but relatively muted Innovator and Ideological tendencies. Qwen2.5:14b delivers the most even coverage across all categories: it peaks in ``Strategic Achievers,” follows with ``Creative Innovators,” and maintains moderate scores in ``Community Support,” ``Emotional,” and the remaining dimensions.

The personality patterns are consistently reflected across different domains. In Subrahmanyan Chandrasekhar's biography (Table~\ref{tab:model_bio}), phi4's strategic achiever orientation leads with career outcomes like `Nobel Prize,' while orca2 emphasizes foundational discoveries (limit discovery ranked first), and qwen2.5 balances achievements with intellectual contributions (`Philosophy of Systematization'). Similarly, in the financial crisis analysis (Table~\ref{tab:financial_crisis_comparison}), phi4 prioritizes immediate catalysts (`Lehman Brothers Bankruptcy'), orca2 incorporates community-wide impacts including `European debt crisis,' while qwen2.5's strategic focus identifies underlying causes (`Subprime Mortgage Crisis') alongside global consequences, demonstrating each model's distinctive values-based decision making as revealed by the personality profiles in Figure~\ref{fig:radar}.

\subsection{Model Semantic Space}

Figure \ref{fig:semantic} positions the three models in a two-dimensional semantic space based on their personality trait embeddings.
The semantic space visualization shows a clear separation between the three models, confirming that they occupy distinct regions in the personality space. Notably, phi4 and qwen2.5.14b appear more distant from each other than either is from orca2.13b, suggesting more contrasting personalities between these two models.
The positioning reveals that phi4 and qwen2.5.14b are both categorized as ``Strategic Achievers" in the semantic space, while orca2.13b stands apart as ``Emotional," reflecting their fundamentally different personality profiles.

\subsection{Scientific Discovery}


\begin{figure}[!h]
    \centering
    \includegraphics[width=0.45\textwidth]{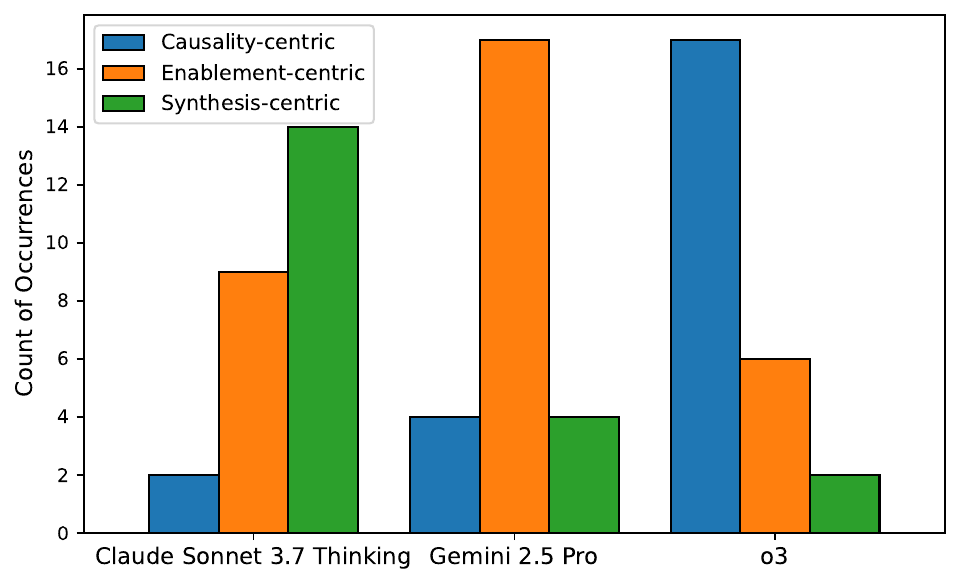}
    \caption{Comparison of reasoning personality profiles across stronger models (Claude Sonnet 3.7, Gemini 2.5 Pro and o3) for the task of critical event sampling and ranking in scientific discoveries.} 
    \label{fig:model_personality_buckets}
\end{figure}

In the scientific discovery category, we characterize model personalities by analyzing how they identify and rank the key events leading to major discoveries. Given the complexity and extended timelines typical of scientific breakthroughs, we focus our evaluation on strong reasoning models: Claude Sonnet 3.7 (with``thinking" enabled), Google Gemini 2.5 Pro, and OpenAI's o3. Also, due to the high cost associated with using these models, we restrict this detailed analysis to the scientific discovery domain only.


We employ a combination of counting key words and open coding to investigate the operational logic of each model. By examining the occurrence of words in the labels, we can identify broad groups of labels centered on causality (e.g., causality, chain, critical), enablement (e.g., enablement, foundation, breakthrough), and synthesis (e.g. conceptual, integration, paradigm). We then use open coding to converge on our final categories after considering the context of the labels and sampled events. As shown in Table \ref{tab:personality_categories_full}, this analysis allows us to identify three distinct guiding principles that reflect the `personality' of each model, namely causality-centric (reflecting a dominant focus on mechanisms and cause-effect pathways, favoring direct cause-and-effect explanations in its sampling), enablement-centric (highlighting the importance of foundations, barrier removal, validation, and making outcomes possible) and synthesis-centric (emphasizing conceptual integration and paradigm-level connections). We use o3 with the finalized three-way codebook to assign each label to the most appropriate category. Our results shown in Fig. \ref{fig:model_personality_buckets} show that o3 and Gemini 2.5 Pro are predominantly causality-centric and enablement-centric, respectively. Claude 3.7 Sonnet is synthesis-centric, but with clear enablement tendencies.



As an example shown in Table~\ref{tab:ml_discoveries_comparison}, the causality focus of o3 is reflected in its top picks: Hopfield’s 1982 energy‑based network paper and the 1986 Nature backpropagation breakthrough, leading to the ``2024 Nobel Prize announcement” that shows how it links discovery directly to outcome. Gemini 2.5 Pro’s enablement tendency is shown in its emphasis on methodological enablers, backpropagation (1986) first, then Hopfield’s associative‑memory application, the Boltzmann machine, and includes `Explicit Use of Physics Principles, highlighting the tools that enable further progress. Claude Sonnet 3.7 shows its balanced personality attributes.



With the growing interest in automating scientific discovery~\citep{o2025sparks}, and human-AI collaboration~\citep{gottweis2025towards}, understanding these reasoning profiles becomes critical.  The current trend of using research papers as datasets for hypothesis generation may be limited; instead, analysing the full timeline (though comparatively difficult to collect) using our proposed scientific discovery dataset could help in better modeling of the multi-dimensional causal chains, hence more accurate hypotheses.

This analysis not only enables more informed model selection, but also points toward designing better human-AI collaboration workflows. By making LLM patterns more interpretable, models can be tasked with solving different tasks, from providing computational scaffolding for complex problems to complementing human expertise, creativity, and values.

Further details, including the sampled events, their rankings, and model-specific personality insights across a range of scientific discoveries, are provided in Section~\ref{sec:apndx}. 

%% file: 7_ablation.tex
\section{Ablation}
\label{sec:ablation}

\begin{table*}[htb]
\centering
\small
\caption{Comparison of models' ranking of critical events in the movie Aladdin}
\label{tab:aladdin_model_comparison}
\begin{tabular}{>{\raggedright\arraybackslash}p{0.31\textwidth}>{\raggedright\arraybackslash}p{0.31\textwidth}>{\raggedright\arraybackslash}p{0.31\textwidth}}
\toprule
\textbf{phi4} & \textbf{orca2:13b} & \textbf{qwen2.5:14b} \\
\midrule
1. Jafar's Plot to Become Sultan & 1. Aladdin meets Jasmine & 1. Aladdin's First Encounter with Jasmine \\
2. Aladdin's Discovery of Cave and Lamp & 2. Aladdin's encounter with Jafar & 2. The Magic Carpet Reveals \\
3. Jasmine Escaping the Palace & 3. Aladdin entering cave and finding lamp & 3. Jafar Unveils His Ambition for Power \\
4. Aladdin's First Wish (Prince Ali) & 4. Aladdin becomes prince & 4. Aladdin Meets Jasmine in Disguise \\
5. The Climactic Battle with Jafar & 5. Final confrontation with Jafar & 5. Sultan Announces Jasmine's Wedding \\
\bottomrule
\end{tabular}
\vspace{2pt}
\caption*{\footnotesize \textbf{Model tendencies:} Phi4 focuses on plot-centric events and villain actions, while Orca2 emphasizes character relationships and transformative moments. Qwen2.5 balances character interactions with narrative developments. All models identify different ``most critical" events, with only the final confrontation appearing consistently (though at different rankings).}
\end{table*}

\subsection{Human-Value Alignment -- Movie Script Analysis}

To examine how different language models prioritize and interpret narrative elements, we utilized the Movie Scripts Dataset from Huggingface, comprising 1,172 movie scripts spanning diverse genres, time periods, and production styles. This dataset provides an ideal testing ground for understanding model-specific values, as humans with different backgrounds and preferences naturally emphasize distinct aspects of the same narrative. By analyzing which events each model identifies as critical within these scripts, we can gain insights into their underlying value systems and interpretative biases.

The movie script analysis validates these personality profiles: phi4's strategic focus is evident in prioritizing business decisions like `Mark Zuckerberg's Decision to Create Facebook' and transformational events such as `Andrew's Practice Regimen'; orca2's emotional orientation highlights relationship conflicts like `Mark's falling out with Eduardo Saverin' and `Andrew's confrontation with Fletcher'; while qwen2.5's balanced yet achievement-oriented approach emphasizes milestone events like `Creation of Facemash' and `The Final Performance at the Lincoln Center' (Tables~\ref{tab:28}--\ref{tab:29}).

This value-driven analysis is exemplified in the rankings of Aladdin movie (Table~\ref{tab:aladdin_model_comparison}), where phi4 prioritizes plot-centric events like `Jafar's Plot to Become Sultan' and `Jasmine Escaping the Palace,' orca2 emphasizes relationship moments such as `Aladdin meets Jasmine' and `Aladdin's encounter with Jafar,' while qwen2.5 balances character interactions (`Aladdin's First Encounter with Jasmine') with narrative developments (`The Magic Carpet Reveals').

Table~\ref{tab:26}--\ref{tab:32} further provides details about the critical events identified by each model across several movies. The above analysis and value signature presented by each model continue to apply to the other movies as well.

These results show that, even with the same script, each model views the story through a different “value lens.” Phi-4 focuses on big strategic or paradigm-shifting moments, Orca2 highlights emotional and relational beats, and Qwen2.5 picks out outcome-driven milestones and clear signs of character agency. Their choices reflect the kinds of narrative biases as found in human readers with different interpretive goals.

%% file: 8_conclusions.tex
\section{Discussion and Conclusion}
\label{sec:conclusion}

Recent work on subjective evaluation and long-context reasoning has underscored the need for benchmarks beyond factual accuracy and needle-in-a-haystack retrieval to probe deeper cognitive functions in LLMs, such as narrative salience detection, value alignment~\citep{meadows2024localvaluebench}, and personality coherence~\citep{jiang2023personallm}. By framing critical-event ranking as a cognitively motivated salience task, the Supernova Event Dataset complements emerging long-context suites like NoLiMa~\citep{modarressi2025nolima} and BABILong~\citep{kuratov2024babilong} while introducing a novel personality-oriented perspective. Our evaluation strategy uses retrieval, structured prompts, and an external LLM as a judge in diverse and extensive scenarios, allowing us to examine model decision making and identify personality patterns in depth.

The dataset and the personality-based evaluation offer a novel way to assess deeper reasoning abilities in large language models. By encouraging models to consider local details and the global context, this task supports better understanding, more thoughtful decision-making, and clearer information organization. It pushes LLMs toward more human-like reasoning rather than surface-level analysis, and it also provides information on how well they align with human values. Our findings align with recent works showing LLMs exhibit personality-like patterns~\citep{jiang2023personallm, bodrovza2024personality}, but extend this by demonstrating that these patterns emerge without explicit personality prompting. This suggests LLMs may have inherent behavioral tendencies shaped by their training, supporting the social determinism view of LLM personality~\citep{yang2024makes}.

In addition to predicting personality traits, future work can use the Supernova Event dataset to evaluate LLMs' ability to model causal chains, understand dynamic event relationships, and perform multi-step reasoning, especially in distinguishing correlation from causation. This will help improve the transparency of their decision-making.


However, several limitations remain. First, while our articles are diverse, they reflect Wikipedia's editorial biases~\citep{greenstein2012wikipedia} and the Western-centric coverage typical of many open corpora~\citep{talat2022you}, which in turn may skew the personality labels inferred by the judge model~\citep{krumdick2025no}.  Second, LLM-as-judge methods are known to exhibit stylistic~\citep{cao2024style} biases that can affect trait inference. One way to mitigate these challenges is to incorporate human annotations and cross-model committees of LLM judges.

While our approach has limitations, particularly the use of LLM-based evaluation without human validation, we believe it opens important avenues for understanding how LLMs approach subjective tasks. The consistent patterns we observe across different domains suggest that LLMs may indeed exhibit stable behavioral tendencies in their decision-making. We invite the community to build upon this work, particularly in developing human validation studies and more rigorous evaluation frameworks. The Supernova Event Dataset and our analysis code are publicly available to facilitate such efforts.

Future versions of the Supernova Event dataset could explore gradient representations of personality traits, incorporate different personality frameworks, and examine how traits change across different narrative contexts and decision scenarios. By integrating mechanistic interpretability techniques to analyze how models internally represent and reason about personality-relevant features during critical event selection, we hope these directions catalyze more transparent, value-aware, and causally grounded large-model research that advances our understanding of AI alignment.

\section{Ethical Considerations}
This work examines the behavioral patterns of large language models (LLMs), which may have implications for their deployment in decision-making contexts. We emphasize that our findings are exploratory and should not be used to make definitive claims about model capabilities. Further validation is needed before using such evaluations for high-stakes decisions.

%% file: 10_appendix.tex
\newpage
\appendix

\section{Appendix}
\label{sec:apndx}

\subsection{Prompts}
\label{sec:prompts}
\begin{promptbox}
You are analyzing the following question: \{question\}

Your task is to extract exactly \textbf{5 critical events} from the provided content. Note that the content may be a biography, a historical event account, or a description of a scientific discovery. Identify key events or milestones that significantly shaped the narrative or understanding of the subject.

Instructions :

- Identify exactly 5 key events or milestones.

- For each event, consider factors such as:

    - Did the event or milestone significantly alter the course or understanding of the subject?
    
    - Was it a turning point that led to a different trajectory?
    
    - If this event had not occurred, would the overall outcome or narrative be drastically different?
    
    - Did the event have long-term consequences that shaped the legacy or interpretation of the subject?
    
- Rank the 5 events from \textbf{most critical to least critical} based on their cascading effects.

- Provide a concise summary of each event in one or at most two sentences.

Return a ranked list of the 5 critical events with their summaries.
\end{promptbox}

\begin{promptbox}
Answer the question based ONLY on the following context:
\{context\}

Question: \{question\}

Please provide your reasoning step-by-step before giving the final answer. In your reasoning:

1. Identify exactly 5 critical events or milestones from the content, which may be a biography, historical event account, or scientific discovery.

2. Rank these events from \textbf{most critical to least critical} based on their impact or cascading effects on the subject.

3. For each event, provide a concise summary in one or two sentences.

4. Explain why the top-ranked event is the most critical, including:

- How it had a cascading effect on the overall narrative or understanding of the subject.

- How the course of events would have been different if this event had not occurred.

- Why this event stands out as the most critical.

Your final answer should include:

- A ranked list of exactly 5 critical events with their summaries.

- A clear statement of the most critical event and its impact.
\end{promptbox}

\begin{promptbox}
You are evaluating the personality of an AI model based purely on the **themes and focus** of the critical events it selects from a movie script. **Do not** classify based on reasoning style (e.g., "Logical", "Analytical", "Methodical"). Instead, focus only on the nature of the events themselves.
        
Below is a response from model "\{model name\}" when asked to identify and rank 5 critical scenes in the event "\{event name\}". **Analyze only the content and themes of the selected events** and classify the personality accordingly.
    
    ---BEGIN MODEL RESPONSE---
    \{full response\}
    ---END MODEL RESPONSE---
    
    Your response should be exactly ONE LINE with just the personality classification into one of the following categories: "Idealogical", "Emotional", "Strategic", "Creative", "Observational", "Public Influence", "Community Support". **Do not** classify based on reasoning style (e.g., "Logical", "Analytical", "Methodical", "Critical")
\end{promptbox}

\subsection{Scientific Discovery Categories}

\begin{table*}[htbp]
\centering
\renewcommand{\arraystretch}{1.2}
\setlength{\tabcolsep}{6pt}
\begin{tabular}{|p{3.2cm}|p{3.8cm}|p{8.8cm}|}
\hline
\textbf{Category} & \textbf{Core Focus} & \textbf{Model Labels} \\ \hline

\textbf{Causality‑centric} &
Mechanisms and cause–effect pathways &
\scriptsize
Breakthrough Chain; Causal Keystones; Causal Links; Causal Pivots; Causal Linchpins; Causal‑Chain Curator; Causal‑Linchpin Lens; Chain‑of‑Causality; Conceptual Causality; Critical Gateways; Critical Path; Critical‑Path; Critical Junctures; Keystone‑Hunter; Mechanism Matters Most; Mechanism-Focused; Mechanistic‑Keystone; Mechanistic Leaps; Mechanistic Proof; Validation Chain; Altered Course
\\ \hline

\textbf{Enablement‑centric} &
Foundational methods that facilitate progress &
\scriptsize
Barrier Breakers; Barrier‑Breaking; Barrier‑Busting; Brake Releasers; Breakthrough \& Validation; Concept Validation; Engineering Enablement; Enabler Focus; Enabling Breakthroughs; Enabling Function; Enabling Impact; Enabling Proof; Essential Enablers; Evidence Centric; Evidentiary Primacy; Experimental Enablement; Foundation \& Enablement; Foundation‑First; Foundational Logic; Foundational Shifts; Foundational Steps; Impact Focus; Impact Sequence; Method Matters; Method Maven; Methodological Innovation; Methodical Innovator; Methodical Revolutionary; Precision Prioritiser; Threshold Focus
\\ \hline

\textbf{Synthesis‑centric} &
Integration and transformation of concepts &
\scriptsize
Adaptive Perseverance; Analogy‑Driven; Analogical Insight; Boundary Transcendence; Concept to Control; Conceptual Integration; Conceptual Necessity; Conceptual Revolution; Discovery to Tool; Framework‑Founding; Gateway Thinking; Interpretive Validation; Mechanistic Convergence; Paradigm Disruption; Paradigm Shifters; Paradigm‑Shift Prioritiser; Predictive Power; Predictive Synergy; Serendipitous Insight; Theory‑Observation Synergy
\\ \hline
\end{tabular}
\caption{Comprehensive mapping of all o3‑generated labels to the three event‑type categories used in our analysis.}
\label{tab:personality_categories_full}
\end{table*}




\begin{table*}[htbp]
\centering
\caption{Scientific Discovery Critical Events Analysis by Model}
\label{tab:1}
\small
\setlength{\tabcolsep}{5pt}
\renewcommand{\arraystretch}{1.3}

\begin{tabular}{|p{1.6cm}|p{14.4cm}|}
\hline
\multicolumn{2}{|c|}{\cellcolor{gray!20}\textbf{Accelerated Expansion of the Universe}} \\
\hline
\multicolumn{2}{|l|}{\cellcolor{gray!10}\textbf{o3}} \\
\hline
\multicolumn{2}{|l|}{\textit{Model Label:} Enabler Focus} \\
\hline
& \quad 1. 1993 Phillips Relation Unveiled \\
& \quad 2. Wide-Field “Batch” Search Devised  \\
& \quad 3. Rival Teams Race Formed \\
& \quad 4. Early-1998 Faintness Result \\
& \quad 5. Back-to-Back Peer-Reviewed Papers (1998–99) \\
\hline

\multicolumn{2}{|l|}{\cellcolor{gray!10}\textbf{Gemini 2.5 Pro}} \\
\hline
\multicolumn{2}{|l|}{\textit{Model Label:} Evidence Centric} \\
\hline
& \quad 1. The Observational Finding of Dimmer Supernovae \\
& \quad 2. Standardization via the Phillips Relationship \\
& \quad 3. Development of Enabling Search and Analysis Techniques\\
& \quad 4. Independent Corroboration by Two Teams \\
& \quad 5. Interpretation of Dimness as Acceleration \\
\hline

\multicolumn{2}{|l|}{\cellcolor{gray!10}\textbf{Claude Sonnet 3.7 Thinking}} \\
\hline
\multicolumn{2}{|l|}{\textit{Model Label:} Precision Prioritiser} \\
\hline
& \quad 1. The Phillips Relationship (1993) \\
& \quad 2. Development of Sophisticated Data Analysis Techniques \\
& \quad 3. Formation of two independent research teams \\
& \quad 4. Development of Systematic Batch Search Strategies  \\
& \quad 5. Access to Premier Astronomical Facilities\\
\hline
\end{tabular}
\end{table*}

\begin{table*}[htbp]
\centering
\caption{Scientific Discovery Critical Events Analysis by Model (continued)}
\label{tab:2}
\small
\setlength{\tabcolsep}{5pt}
\renewcommand{\arraystretch}{1.3}

\begin{tabular}{|p{1.6cm}|p{14.4cm}|}
\hline
\multicolumn{2}{|c|}{\cellcolor{gray!20}\textbf{Foundational Discoveries Enabling Machine Learning with Artificial Neural Networks}} \\
\hline
\multicolumn{2}{|l|}{\cellcolor{gray!10}\textbf{o3}} \\
\hline
\multicolumn{2}{|l|}{\textit{Model Label:} Chain-of-Causality} \\
\hline
& \quad 1. Hopfield’s 1982 energy-based network paper \\
& \quad 2. 1986 Nature backpropagation breakthrough \\
& \quad 3. 1985 Boltzmann-machine formulation \\
& \quad 4. 1969 Perceptrons critique and ensuing AI Winter \\
& \quad 5. 2024 Nobel Prize announcement  \\
\hline

\multicolumn{2}{|l|}{\cellcolor{gray!10}\textbf{Gemini 2.5 Pro}} \\
\hline
\multicolumn{2}{|l|}{\textit{Model Label:} Enabling Breakthroughs} \\
\hline
& \quad 1. The Popularization of Backpropagation (1986 Nature Paper) \\
& \quad 2. Hopfield's Application of Physics to Associative Memory (1982 PNAS Paper) \\
& \quad 3. The Development of the Boltzmann Machine (1985 Cognitive Science Paper) \\
& \quad 4. Minsky and Papert's "Perceptrons" Critique (1969) \\
& \quad 5. The Explicit Use of Physics Principles \\
\hline

\multicolumn{2}{|l|}{\cellcolor{gray!10}\textbf{Claude Sonnet 3.7 Thinking}} \\
\hline
\multicolumn{2}{|l|}{\textit{Model Label:} Barrier Breakers} \\
\hline
& \quad 1. Backpropagation Paper in Nature (1986) \\
& \quad 2. Hopfield's Network Model in PNAS (1982) \\
& \quad 3. Minsky and Papert's Perceptrons Book (1969) \\
& \quad 4. Boltzmann Machine Paper (1985) \\
& \quad 5. Rosenblatt's Perceptron (late 1950s) \\
\hline
\end{tabular}
\end{table*}

\begin{table*}[htbp]
\centering
\caption{Scientific Discovery Critical Events Analysis by Model (continued)}
\label{tab:3}
\small
\setlength{\tabcolsep}{5pt}
\renewcommand{\arraystretch}{1.3}

\begin{tabular}{|p{1.6cm}|p{14.4cm}|}
\hline
\multicolumn{2}{|c|}{\cellcolor{gray!20}\textbf{Black Hole Formation and the Galactic Center Supermassive Object}} \\
\hline
\multicolumn{2}{|l|}{\cellcolor{gray!10}\textbf{o3}} \\
\hline
\multicolumn{2}{|l|}{\textit{Model Label:} Causal Keystones} \\
\hline
& \quad 1. Penrose’s 1965 singularity theorem introduces “trapped surfaces.” \\
& \quad 2. 1974 discovery of the compact radio source Sagittarius A* \\
& \quad 3. Early-1990s launch of long-term infrared speckle-imaging campaigns by Genzel and Ghez teams \\
& \quad 4. Late-1990s deployment of adaptive-optics systems on large telescopes \\
& \quad 5. 2002–2003 completion of star S2’s 16-year orbit around Sgr A* \\
\hline

\multicolumn{2}{|l|}{\cellcolor{gray!10}\textbf{Gemini 2.5 Pro}} \\
\hline
\multicolumn{2}{|l|}{\textit{Model Label:} Foundation \& Enablement} \\
\hline
& \quad 1. Penrose's Singularity Theorem (1965) \\
& \quad 2. Development and Application of Adaptive Optics (AO) \\
& \quad 3. Mapping of Star S2/S0-2's Complete Orbit (culminating ~2002-2003) \\
& \quad 4. Development and Application of Speckle Imaging \\
& \quad 5. Independent Confirmation by Genzel and Ghez Teams \\
\hline

\multicolumn{2}{|l|}{\cellcolor{gray!10}\textbf{Claude Sonnet 3.7 Thinking}} \\
\hline
\multicolumn{2}{|l|}{\textit{Model Label:} Theory-Observation Synergy} \\
\hline
& \quad 1. Penrose's 1965 Trapped Surface Concept (1965) \\
& \quad 2. Development of Advanced Infrared Observation Techniques (1990s) \\
& \quad 3. Formation of Competing Observational Teams (Early 1990s) \\
& \quad 4. Mapping of Star S2/S0-2's Complete Orbit (2002-2003) \\
& \quad 5. Detection of Relativistic Effects in Stellar Orbits (2018-2020) \\
\hline
\end{tabular}
\end{table*}

\begin{table*}[htbp]
\centering
\caption{Scientific Discovery Critical Events Analysis by Model (continued)}
\label{tab:4}
\small
\setlength{\tabcolsep}{5pt}
\renewcommand{\arraystretch}{1.3}

\begin{tabular}{|p{1.6cm}|p{14.4cm}|}
\hline
\multicolumn{2}{|c|}{\cellcolor{gray!20}\textbf{X-rays}} \\
\hline
\multicolumn{2}{|l|}{\cellcolor{gray!10}\textbf{o3}} \\
\hline
\multicolumn{2}{|l|}{\textit{Model Label:} Causal-Chain Curator} \\
\hline
& \quad 1. 1895-11-08 – Röntgen sees a baffling green glow from a shielded Crookes tube \\
& \quad 2. 1875-1879 – Crookes perfects the high-vacuum discharge tube \\
& \quad 3. 1895-12-22 – First human radiograph (Bertha Röntgen’s hand) \\
& \quad 4. 1895-12-28 → early 1896 – Lightning-fast publication and photo mail-out \\
& \quad 5. 1892-1894 – Lenard’s thin-window cathode-ray experiments \\
\hline

\multicolumn{2}{|l|}{\cellcolor{gray!10}\textbf{Gemini 2.5 Pro}} \\
\hline
\multicolumn{2}{|l|}{\textit{Model Label:} Impact Sequence} \\
\hline
& \quad 1. Röntgen's Accidental Observation (November 8, 1895) \\
& \quad 2. Röntgen's Systematic Characterization \\
& \quad 3. The X-ray Photograph of Bertha Röntgen's Hand \\
& \quad 4. Rapid Publication and Direct Mailing of Findings (Late Dec 1895 / Early Jan 1896)  \\
& \quad 5. Award of the First Nobel Prize in Physics (1901) \\
\hline

\multicolumn{2}{|l|}{\cellcolor{gray!10}\textbf{Claude Sonnet 3.7 Thinking}} \\
\hline
\multicolumn{2}{|l|}{\textit{Model Label:} Enabling Impact} \\
\hline
& \quad 1. Development of improved vacuum technology (1855-1865) \\
& \quad 2. Röntgen's initial observation (November 8, 1895) \\
& \quad 3. Röntgen's systematic investigation (November-December 1895) \\
& \quad 4. Creation of first "röntgenogram" (December 22, 1895) \\
& \quad 5. Publication and dissemination (December 28, 1895 - January 1896) \\
\hline
\end{tabular}
\end{table*}

\begin{table*}[htbp]
\centering
\caption{Scientific Discovery Critical Events Analysis by Model (continued)}
\label{tab:5}
\small
\setlength{\tabcolsep}{5pt}
\renewcommand{\arraystretch}{1.3}

\begin{tabular}{|p{1.6cm}|p{14.4cm}|}
\hline
\multicolumn{2}{|c|}{\cellcolor{gray!20}\textbf{Low-Loss Optical Fibers and Charge-Coupled Device Sensors}} \\
\hline
\multicolumn{2}{|l|}{\cellcolor{gray!10}\textbf{o3}} \\
\hline
\multicolumn{2}{|l|}{\textit{Model Label:} Threshold Focus} \\
\hline
& \quad 1. Kao-Hockham Purity Hypothesis \& 20 dB/km Target (1966)  \\
& \quad 2. Boyle \& Smith Conceive the Charge-Coupled Device (Oct 1969)  \\
& \quad 3. Corning’s 17 dB/km Fiber – “Whoopee!” (Aug 1970) \\
& \quad 4. First Laser Produces Coherent Light (May 16 1960) \\
& \quad 5. Tompsett Turns the CCD into a Camera (1970-1973)  \\
\hline

\multicolumn{2}{|l|}{\cellcolor{gray!10}\textbf{Gemini 2.5 Pro}} \\
\hline
\multicolumn{2}{|l|}{\textit{Model Label:}  Foundational Shifts} \\
\hline
& \quad 1. Kao's Impurity Hypothesis (Early 1960s) \\
& \quad 2. Corning's 1970 Low-Loss Fiber Demonstration \\
& \quad 3. Boyle \& Smith's CCD Conception (October 1969) \\
& \quad 4. Kao \& Hockham's 1966 Paper \\
& \quad 5. Recognition and Pursuit of CCD Imaging Potential (Post-1969) \\
\hline

\multicolumn{2}{|l|}{\cellcolor{gray!10}\textbf{Claude Sonnet 3.7 Thinking}} \\
\hline
\multicolumn{2}{|l|}{\textit{Model Label:} Conceptual Revolution} \\
\hline
& \quad 1. Kao's impurity hypothesis (mid-1960s) \\
& \quad 2. Boyle-Smith CCD conception (October 1969) \\
& \quad 3. Corning breaks 20 dB/km barrier (August 1970)\\
& \quad 4. CCD pivot to imaging applications (1969-1970) \\
& \quad 5. Kao-Hockham paper publication (1966) \\
\hline
\end{tabular}
\end{table*}

\begin{table*}[htbp]
\centering
\caption{Scientific Discovery Critical Events Analysis by Model (continued)}
\label{tab:6}
\small
\setlength{\tabcolsep}{5pt}
\renewcommand{\arraystretch}{1.3}

\begin{tabular}{|p{1.6cm}|p{14.4cm}|}
\hline
\multicolumn{2}{|c|}{\cellcolor{gray!20}\textbf{Computational Protein Design and Structure Prediction}} \\
\hline
\multicolumn{2}{|l|}{\cellcolor{gray!10}\textbf{o3}} \\
\hline
\multicolumn{2}{|l|}{\textit{Model Label:} Critical-Path} \\
\hline
& \quad 1. AlphaFold2 triumph at CASP14 (Nov 2020)  \\
& \quad 2. Top7 de novo design confirmation (2003) \\
& \quad 3. Launch of the CASP blind benchmark (1994) \\
& \quad 4. Anfinsen’s refolding experiments \& “sequence-codes-structure” dogma (1950s–60s) \\
& \quad 5. Open-source release of AlphaFold2 and the 200 M-entry AlphaFold DB (July 2021) \\
\hline

\multicolumn{2}{|l|}{\cellcolor{gray!10}\textbf{Gemini 2.5 Pro}} \\
\hline
\multicolumn{2}{|l|}{\textit{Model Label:} Breakthrough \& Validation} \\
\hline
& \quad 1. The Design and Experimental Validation of Top7 (2003) \\
& \quad 2. The Architectural Redesign Leading to AlphaFold2 (post-CASP13, ~2018-2020) \\
& \quad 3. AlphaFold2's Breakthrough Performance at CASP14 (2020) \\
& \quad 4. The Conceptual Leap to Invert Prediction for Design (mid-to-late 1990s) \\
& \quad 5. Establishment of the CASP Experiment (1994) \\
\hline

\multicolumn{2}{|l|}{\cellcolor{gray!10}\textbf{Claude Sonnet 3.7 Thinking}} \\
\hline
\multicolumn{2}{|l|}{\textit{Model Label:} Paradigm Shifters} \\
\hline
& \quad 1. Development of AlphaFold2's architecture (2018-2020)  \\
& \quad 2. Design of Top7 by Baker's lab (2003) \\
& \quad 3. AlphaFold2's performance at CASP14 (2020) \\
& \quad 4. Development of the Rosetta software suite (1990s-2000s)  \\
& \quad 5. Creation of the CASP competition (1994)  \\
\hline
\end{tabular}
\end{table*}

\begin{table*}[htbp]
\centering
\caption{Scientific Discovery Critical Events Analysis by Model (continued)}
\label{tab:7}
\small
\setlength{\tabcolsep}{5pt}
\renewcommand{\arraystretch}{1.3}

\begin{tabular}{|p{1.6cm}|p{14.4cm}|}
\hline
\multicolumn{2}{|c|}{\cellcolor{gray!20}\textbf{Radioactivity}} \\
\hline
\multicolumn{2}{|l|}{\cellcolor{gray!10}\textbf{o3}} \\
\hline
\multicolumn{2}{|l|}{\textit{Model Label:} Critical Path} \\
\hline
& \quad 1. Becquerel’s dark-drawer revelation (Feb 26 – Mar 1 1896) \\
& \quad 2. Röntgen’s discovery of X-rays (Nov 8 1895) \\
& \quad 3. Marie Curie’s quantitative electrometer studies \& atomic-property hypothesis (1897 – Apr 1898) \\
& \quad 4. Curies’ radiochemical discovery and concentration of radium (Dec 1898 → 1902)\\
& \quad 5. 1903 Nobel Prize shared by Becquerel and the Curies (Dec 10 1903) \\
\hline

\multicolumn{2}{|l|}{\cellcolor{gray!10}\textbf{Gemini 2.5 Pro}} \\
\hline
\multicolumn{2}{|l|}{\textit{Model Label:} Altered Course} \\
\hline
& \quad 1. Becquerel's Serendipitous Discovery of Spontaneous Emission (Feb 26 - Mar 1, 1896) \\
& \quad 2. Marie Curie's Hypothesis of Radioactivity as an Atomic Property (Apr 1898)\\
& \quad 3. Marie Curie Begins Systematic Study Using Electrometer (1897) \\
& \quad 4. P. \& M. Curie Announce Discovery of Radium (Dec 1898)\\
& \quad 5. Poincaré's Suggestion Linking X-rays and Fluorescence (Jan 1896) \\
\hline

\multicolumn{2}{|l|}{\cellcolor{gray!10}\textbf{Claude Sonnet 3.7 Thinking}} \\
\hline
\multicolumn{2}{|l|}{\textit{Model Label:} Serendipitous Insight} \\
\hline
& \quad 1. Becquerel's unexpected discovery (March 1-2, 1896) \\
& \quad 2. Röntgen's discovery of X-rays (November 1895) \\
& \quad 3. Cloudy weather in Paris (February 26-March 1, 1896) \\
& \quad 4. Marie Curie's atomic property hypothesis (April 1898)\\
& \quad 5. Marie Curie's systematic study using the electrometer (1897) \\
\hline
\end{tabular}
\end{table*}

\begin{table*}[htbp]
\centering
\caption{Scientific Discovery Critical Events Analysis by Model (continued)}
\label{tab:8}
\small
\setlength{\tabcolsep}{5pt}
\renewcommand{\arraystretch}{1.3}

\begin{tabular}{|p{1.6cm}|p{14.4cm}|}
\hline
\multicolumn{2}{|c|}{\cellcolor{gray!20}\textbf{Stellar Structure, Evolution, and Nucleosynthesis}} \\
\hline
\multicolumn{2}{|l|}{\cellcolor{gray!10}\textbf{o3}} \\
\hline
\multicolumn{2}{|l|}{\textit{Model Label:} Critical Gateways} \\
\hline
& \quad 1. Chandrasekhar derives the white-dwarf mass limit (1930-35) – \\
& \quad 2. Carbon-12 resonance predicted and confirmed (1953)  \\
& \quad 3. B²FH synthesis published (1957) \\
& \quad 4. Fowler’s post-war accelerator program (late-1940s–mid-1950s) \\
& \quad 5. Pulsars discovered as neutron stars (1967) \\
\hline

\multicolumn{2}{|l|}{\cellcolor{gray!10}\textbf{Gemini 2.5 Pro}} \\
\hline
\multicolumn{2}{|l|}{\textit{Model Label:} Foundational Steps} \\
\hline
& \quad 1. Chandrasekhar's Derivation of the Mass Limit (1930-1935) \\
& \quad 2. Publication of the B²FH Paper (1957) \\
& \quad 3. Experimental Confirmation of the Carbon-12 Resonant State (1953) \\
& \quad 4. Fowler's Initiation of the Experimental Nuclear Astrophysics Program (late 1930s onward)
 \\
& \quad 5. Eddington's Public Rejection of the Chandrasekhar Limit (1935) \\
\hline

\multicolumn{2}{|l|}{\cellcolor{gray!10}\textbf{Claude Sonnet 3.7 Thinking}} \\
\hline
\multicolumn{2}{|l|}{\textit{Model Label:} Predictive Synergy} \\
\hline
& \quad 1. Chandrasekhar's relativistic insight (1930-1931) \\
& \quad 2. Fowler's experimental program (late 1930s) \\
& \quad 3. Hoyle's Carbon-12 prediction (1953) \\
& \quad 4. Experimental confirmation of Hoyle's prediction (1953) \\
& \quad 5. The B²FH paper (1957) \\
\hline
\end{tabular}
\end{table*}

\begin{table*}[htbp]
\centering
\caption{Scientific Discovery Critical Events Analysis by Model (continued)}
\label{tab:9}
\small
\setlength{\tabcolsep}{5pt}
\renewcommand{\arraystretch}{1.3}

\begin{tabular}{|p{1.6cm}|p{14.4cm}|}
\hline
\multicolumn{2}{|c|}{\cellcolor{gray!20}\textbf{Zeeman Effect}} \\
\hline
\multicolumn{2}{|l|}{\cellcolor{gray!10}\textbf{o3}} \\
\hline
\multicolumn{2}{|l|}{\textit{Model Label:} Validation Chain} \\
\hline
& \quad 1. Sept 2 1896 – Zeeman’s first broadening of the sodium D-lines \\
& \quad 2. Nov 2 1896 – Lorentz’s lightning-fast electron-theory interpretation \\
& \quad 3. Early 1897 – Experimental resolution of triplet/doublet and polarization \\
& \quad 4. Sept–Oct 1896 – Rigorous control experiments to rule out flame artifacts \\
& \quad 5. Feb–Mar 1897 – Swift publication and global replication \\
\hline

\multicolumn{2}{|l|}{\cellcolor{gray!10}\textbf{Gemini 2.5 Pro}} \\
\hline
\multicolumn{2}{|l|}{\textit{Model Label:} Interpretive Validation} \\
\hline
& \quad 1. Lorentz's Swift Theoretical Explanation (Nov 1896) \\
& \quad 2. Zeeman's Initial Observation of Line Broadening (Sept 1896) \\
& \quad 3. Zeeman's Confirmation of Splitting and Polarization (Early 1897)  \\
& \quad 4. Preston \& Michelson's Discovery of Anomalous Splitting (1897-1898) \\
& \quad 5. Zeeman's Careful Elimination of Experimental Artifacts (Sept-Oct 1896)\\
\hline

\multicolumn{2}{|l|}{\cellcolor{gray!10}\textbf{Claude Sonnet 3.7 Thinking}} \\
\hline
\multicolumn{2}{|l|}{\textit{Model Label:} Predictive Power} \\
\hline
& \quad 1. Zeeman's Initial Observation (September 2, 1896) \\
& \quad 2. Lorentz's Theoretical Interpretation (November 2, 1896) \\
& \quad 3. Confirmation of Line Splitting and Polarization (Early 1897) \\
& \quad 4. Discovery of "Anomalous" Zeeman Effect (1897-1898) \\
& \quad 5. Explanation via Quantum Mechanics and Electron Spin (1925) \\
\hline
\end{tabular}
\end{table*}

\begin{table*}[htbp]
\centering
\caption{Scientific Discovery Critical Events Analysis by Model (continued)}
\label{tab:10}
\small
\setlength{\tabcolsep}{5pt}
\renewcommand{\arraystretch}{1.3}

\begin{tabular}{|p{1.6cm}|p{14.4cm}|}
\hline
\multicolumn{2}{|c|}{\cellcolor{gray!20}\textbf{Catalytic RNA}} \\
\hline
\multicolumn{2}{|l|}{\cellcolor{gray!10}\textbf{o3}} \\
\hline
\multicolumn{2}{|l|}{\textit{Model Label:} Paradigm-Shift Prioritiser} \\
\hline
& \quad 1. 1981 – The “minus-extract” epiphany \\
& \quad 2. 1983 – RNA-only RNase P catalysis \\
& \quad 3. 1978 – RNase P’s essential RNA subunit revealed \\
& \quad 4. 1982/84 – Synthetic RNA passes the purity test \\
& \quad 5. 1982 – “Ribozyme” coined \& cross-system validation \\
\hline

\multicolumn{2}{|l|}{\cellcolor{gray!10}\textbf{Gemini 2.5 Pro}} \\
\hline
\multicolumn{2}{|l|}{\textit{Model Label:} Evidentiary Primacy} \\
\hline
& \quad 1. M1 RNA proven catalytic (Altman, 1983)  \\
& \quad 2. Control reveals self-splicing (Cech, ~1981) \\
& \quad 3. Synthetic RNA self-splices (Cech, 1982)  \\
& \quad 4. Synthetic M1 RNA confirms catalysis (Altman, 1984)  \\
& \quad 5. Essential RNA in enzyme identified (Altman, 1978) \\
\hline

\multicolumn{2}{|l|}{\cellcolor{gray!10}\textbf{Claude Sonnet 3.7 Thinking}} \\
\hline
\multicolumn{2}{|l|}{\textit{Model Label:} Paradigm Disruption} \\
\hline
& \quad 1. Unexpected Self-Splicing in Control Experiment (Cech Lab) \\
& \quad 2. M1 RNA Catalysis Under High-Salt Conditions (Altman Lab) \\
& \quad 3. Synthetic Pre-rRNA Self-Splicing Confirmation (Cech Lab) \\
& \quad 4. Synthetic M1 RNA Retaining Catalytic Activity (Altman Lab) \\
& \quad 5. Synthetic M1 RNA Retaining Catalytic Activity (Altman Lab) \\
\hline
\end{tabular}
\end{table*}

\begin{table*}[htbp]
\centering
\caption{Scientific Discovery Critical Events Analysis by Model (continued)}
\label{tab:11}
\small
\setlength{\tabcolsep}{5pt}
\renewcommand{\arraystretch}{1.3}

\begin{tabular}{|p{1.6cm}|p{14.4cm}|}
\hline
\multicolumn{2}{|c|}{\cellcolor{gray!20}\textbf{Quantum Dots}} \\
\hline
\multicolumn{2}{|l|}{\cellcolor{gray!10}\textbf{o3}} \\
\hline
\multicolumn{2}{|l|}{\textit{Model Label:} Causal Links} \\
\hline
& \quad 1. 1993 — Hot-Injection Synthesis Perfected \\
& \quad 2. 1983-84 — Brus Colloids \& Brus Equation \\
& \quad 3. 1981 — Yekimov Glass Experiments \\
& \quad 4. 1982 — Efros Quantum-Confinement Theory \\
& \quad 5. 2023 — Nobel Prize Awarded \\
\hline

\multicolumn{2}{|l|}{\cellcolor{gray!10}\textbf{Gemini 2.5 Pro}} \\
\hline
\multicolumn{2}{|l|}{\textit{Model Label:} Impact Focus} \\
\hline
& \quad 1. Bawendi's Hot-Injection Synthesis (1993)  \\
& \quad 2. Yekimov's Controlled Creation \& Observation (~1981)  \\
& \quad 3. Brus's Independent Discovery \& Colloidal System (~1983)  \\
& \quad 4. Fröhlich's Theoretical Prediction (1937)  \\
& \quad 5. Differential Publication Impact (Early 1980s) \\
\hline

\multicolumn{2}{|l|}{\cellcolor{gray!10}\textbf{Claude Sonnet 3.7 Thinking}} \\
\hline
\multicolumn{2}{|l|}{\textit{Model Label:} Concept to Control} \\
\hline
& \quad 1. Yekimov's Experimental Demonstration (1981) \\
& \quad 2. Bawendi's "Hot-Injection" Synthesis Method (1993) \\
& \quad 3. Brus's Independent Discovery in Colloids (1982-1983) \\
& \quad 4. Brus's Theoretical Framework (1983-1984) \\
& \quad 5. First Commercial Applications (Early 2000s) \\
\hline
\end{tabular}
\end{table*}

\begin{table*}[htbp]
\centering
\caption{Scientific Discovery Critical Events Analysis by Model (continued)}
\label{tab:12}
\small
\setlength{\tabcolsep}{5pt}
\renewcommand{\arraystretch}{1.3}

\begin{tabular}{|p{1.6cm}|p{14.4cm}|}
\hline
\multicolumn{2}{|c|}{\cellcolor{gray!20}\textbf{Insulin}} \\
\hline
\multicolumn{2}{|l|}{\cellcolor{gray!10}\textbf{o3}} \\
\hline
\multicolumn{2}{|l|}{\textit{Model Label:} Method Maven} \\
\hline
& \quad 1. FDNB N-terminal labeling invented (1945)  \\
& \quad 2. Fragmentation-and-overlap strategy conceived (~1946-48)   \\
& \quad 3. Two-dimensional peptide “fingerprinting” perfected (late-1940s)   \\
& \quad 4. Insulin chosen as the target (1943-44) \\
& \quad 5. Disulfide bridges mapped \& full structure published (1955) – \\
\hline

\multicolumn{2}{|l|}{\cellcolor{gray!10}\textbf{Gemini 2.5 Pro}} \\
\hline
\multicolumn{2}{|l|}{\textit{Model Label:} Foundational Logic} \\
\hline
& \quad 1. FDNB N-Terminal Analysis  \\
& \quad 2. Fragmentation and Overlap Principle \\
& \quad 3. Development of Peptide Fingerprinting \\
& \quad 4. Publication and Conceptual Breakthrough \\
& \quad 5. Disulfide Bond Mapping \\
\hline

\multicolumn{2}{|l|}{\cellcolor{gray!10}\textbf{Claude Sonnet 3.7 Thinking}} \\
\hline
\multicolumn{2}{|l|}{\textit{Model Label:} Method Matters} \\
\hline
& \quad 1. Development of the FDNB (N-Terminal) Method (c. 1945) \\
& \quad 2. Conceptualization of the "Overlap Principle" Strategy \\
& \quad 3. Development of 2D Paper Chromatography ("Fingerprinting")  \\
& \quad 4. Strategic Selection of Insulin as Target Protein (1943-1944) \\
& \quad 5. Determination of Disulfide Bridge Locations (1955) \\
\hline
\end{tabular}
\end{table*}

\begin{table*}[htbp]
\centering
\caption{Scientific Discovery Critical Events Analysis by Model (continued)}
\label{tab:13}
\small
\setlength{\tabcolsep}{5pt}
\renewcommand{\arraystretch}{1.3}

\begin{tabular}{|p{1.6cm}|p{14.4cm}|}
\hline
\multicolumn{2}{|c|}{\cellcolor{gray!20}\textbf{Spontaneous Symmetry Breaking and Quark Generations}} \\
\hline
\multicolumn{2}{|l|}{\cellcolor{gray!10}\textbf{o3}} \\
\hline
\multicolumn{2}{|l|}{\textit{Model Label:} Causal linchpins} \\
\hline
& \quad 1. Nambu’s 1960-61 leap to Spontaneous Symmetry Breaking (SSB)  \\
& \quad 2. 1964 Higgs-Englert-Brout mechanism \\
& \quad 3. Cronin–Fitch 1964 discovery of CP violation in kaons \\
& \quad 4. Kobayashi–Maskawa 1973 prediction of a third quark generation \\
& \quad 5. 1977 discovery of the bottom (b) quark \\
\hline

\multicolumn{2}{|l|}{\cellcolor{gray!10}\textbf{Gemini 2.5 Pro}} \\
\hline
\multicolumn{2}{|l|}{\textit{Model Label:} Conceptual Necessity} \\
\hline
& \quad 1. Nambu's Application of SSB Analogy (1960-1961):  \\
& \quad 2. Kobayashi \& Maskawa's 6-Quark / CKM Proposal (1973)  \\
& \quad 3. KM's Proof of 4-Quark Insufficiency (1972) \\
& \quad 4. Cronin \& Fitch Discovery of CP Violation (1964) \\
& \quad 5. Higgs Mechanism Proposal (1964) \\
\hline

\multicolumn{2}{|l|}{\cellcolor{gray!10}\textbf{Claude Sonnet 3.7 Thinking}} \\
\hline
\multicolumn{2}{|l|}{\textit{Model Label:} Conceptual Causality} \\
\hline
& \quad 1. Nambu's Application of Superconductivity to Particle Physics (1960-1961) \\
& \quad 2. Discovery of CP Violation by Cronin and Fitch (1964)
 \\
& \quad 3. Kobayashi and Maskawa's Six-Quark Model Proposal (1973)
 \\
& \quad 4. Discovery of the Bottom Quark (1977) \\
& \quad 5. B-Factory Experiments Confirming CP Violation in B Mesons (c. 2001)
 \\
\hline
\end{tabular}
\end{table*}

\begin{table*}[htbp]
\centering
\caption{Scientific Discovery Critical Events Analysis by Model (continued)}
\label{tab:14}
\small
\setlength{\tabcolsep}{5pt}
\renewcommand{\arraystretch}{1.3}

\begin{tabular}{|p{1.6cm}|p{14.4cm}|}
\hline
\multicolumn{2}{|c|}{\cellcolor{gray!20}\textbf{Cancer Therapy by Inhibition of Negative Immune Regulation}} \\
\hline
\multicolumn{2}{|l|}{\cellcolor{gray!10}\textbf{o3}} \\
\hline
\multicolumn{2}{|l|}{\textit{Model Label:} Mechanistic Proof} \\
\hline
& \quad 1. 1995 – CTLA-4 proven an immune “brake” \\
& \quad 2. 1996 – Tumor rejection after CTLA-4 blockade in mice \\
& \quad 3. 1999 – PD-1 knockout mice reveal a second checkpoint \\
& \quad 4. 2000 – PD-L1 identified as PD-1’s ligand \\
& \quad 5. 2011 – FDA approval of ipilimumab (anti-CTLA-4) \\
\hline

\multicolumn{2}{|l|}{\cellcolor{gray!10}\textbf{Gemini 2.5 Pro}} \\
\hline
\multicolumn{2}{|l|}{\textit{Model Label:} Concept validation} \\
\hline
& \quad 1. Allison Lab Demonstrates Anti-CTLA-4 Antibody Shrinks Tumors in Mice (1996)  \\
& \quad 2. Allison Lab Determines CTLA-4's Inhibitory Function (1995) \\
& \quad 3. Honjo Lab Shows PD-1 Knockout Mice Develop Autoimmune Disease (1999) \\
& \quad 4. Honjo Lab Discovers the PD-1 Gene (1992) \\
& \quad 5. Honjo Lab/Collaborators Identify PD-L1 and Demonstrate PD-1/PD-L1 Inhibition (2000) \\
\hline

\multicolumn{2}{|l|}{\cellcolor{gray!10}\textbf{Claude Sonnet 3.7 Thinking}} \\
\hline
\multicolumn{2}{|l|}{\textit{Model Label:} Brake releasers} \\
\hline
& \quad 1. Anti-CTLA-4 Antibody Blocks Tumor Growth in Mice (1996) \\
& \quad 2. Discovery of CTLA-4's Inhibitory Function (1995) \\
& \quad 3. PD-1 Knockout Mice Reveal Its Role as Immune Brake (1999) \\
& \quad 4. FDA Approval of Ipilimumab (2011)  \\
& \quad 5. Identification of PD-L1 as PD-1's Ligand (2000) \\
\hline
\end{tabular}
\end{table*}

\begin{table*}[htbp]
\centering
\caption{Scientific Discovery Critical Events Analysis by Model (continued)}
\label{tab:15}
\small
\setlength{\tabcolsep}{5pt}
\renewcommand{\arraystretch}{1.3}

\begin{tabular}{|p{1.6cm}|p{14.4cm}|}
\hline
\multicolumn{2}{|c|}{\cellcolor{gray!20}\textbf{Antiproton}} \\
\hline
\multicolumn{2}{|l|}{\cellcolor{gray!10}\textbf{o3}} \\
\hline
\multicolumn{2}{|l|}{\textit{Model Label:} Critical-Path} \\
\hline
& \quad 1. Bevatron completed (1954)  \\
& \quad 2. Segrè-Chamberlain coincidence detection scheme (1955)  \\
& \quad 3. Dirac’s antiparticle prediction (1931) \\
& \quad 4. Anderson discovers the positron (1932)  \\
& \quad 5. Emulsion “star” confirms annihilation (late 1955) \\
\hline

\multicolumn{2}{|l|}{\cellcolor{gray!10}\textbf{Gemini 2.5 Pro}} \\
\hline
\multicolumn{2}{|l|}{\textit{Model Label:} Experimental Enablement} \\
\hline
& \quad 1. Bevatron Completion (1954) \\
& \quad 2. Development of Detection Methodology (1955) \\
& \quad 3. Anderson's Positron Discovery (1932) \\
& \quad 4. Dirac's Positron Prediction (1931) \\
& \quad 5. Successful Counter Experiment Run (Sept 1955)  \\
\hline

\multicolumn{2}{|l|}{\cellcolor{gray!10}\textbf{Claude Sonnet 3.7 Thinking}} \\
\hline
\multicolumn{2}{|l|}{\textit{Model Label:} Breakthrough Chain} \\
\hline
& \quad 1. Dirac's Relativistic Equation (1928) \\
& \quad 2. Anderson's Discovery of the Positron (1932) \\
& \quad 3. Construction of the Bevatron (1947-1954) \\
& \quad 4. The Sophisticated Detection System Design \\
& \quad 5. Visual Confirmation in Photographic Emulsions \\
\hline
\end{tabular}
\end{table*}

\begin{table*}[htbp]
\centering
\caption{Scientific Discovery Critical Events Analysis by Model (continued)}
\label{tab:16}
\small
\setlength{\tabcolsep}{5pt}
\renewcommand{\arraystretch}{1.3}

\begin{tabular}{|p{1.6cm}|p{14.4cm}|}
\hline
\multicolumn{2}{|c|}{\cellcolor{gray!20}\textbf{Interpretation of the Genetic Code}} \\
\hline
\multicolumn{2}{|l|}{\cellcolor{gray!10}\textbf{o3}} \\
\hline
\multicolumn{2}{|l|}{\textit{Model Label:} Barrier-Busting} \\
\hline
& \quad 1. Frameshift Triplet Proof (Crick \& Brenner, 1961)  \\
& \quad 2. Poly-U Cell-Free Breakthrough (Nirenberg \& Matthaei, May 1961) \\
& \quad 3. Trinucleotide Binding Assay (Nirenberg \& Leder, 1964)  \\
& \quad 4. Defined Polymer Synthesis (Khorana, 1963-1966)  \\
& \quad 5. First tRNA Sequence (Holley, 1965)  \\
\hline

\multicolumn{2}{|l|}{\cellcolor{gray!10}\textbf{Gemini 2.5 Pro}} \\
\hline
\multicolumn{2}{|l|}{\textit{Model Label:} Mechanism-Focused} \\
\hline
& \quad 1. Nirenberg \& Matthaei's Poly-U Experiment (1961)  \\
& \quad 2. Nirenberg \& Leder's Trinucleotide Binding Assay (1964)  \\
& \quad 3. Khorana's Synthesis of Defined Repeating Polynucleotides  \\
& \quad 4. Holley's Sequencing of Alanine tRNA (1965)  \\
& \quad 5. Crick, Brenner et al.'s Frameshift Mutation Experiments (1961)  \\
\hline

\multicolumn{2}{|l|}{\cellcolor{gray!10}\textbf{Claude Sonnet 3.7 Thinking}} \\
\hline
\multicolumn{2}{|l|}{\textit{Model Label:} Methodological Innovation} \\
\hline
& \quad 1. Nirenberg and Matthaei's Poly-U Experiment (May 1961)  \\
& \quad 2. Nirenberg-Leder Trinucleotide Binding Assay (1964) \\
& \quad 3. Khorana's Synthetic Polynucleotides  \\
& \quad 4. Discovery of Transfer RNA (late 1950s) \\
& \quad 5. Holley's Alanine tRNA Sequencing (1965) \\
\hline
\end{tabular}
\end{table*}

\begin{table*}[htbp]
\centering
\caption{Scientific Discovery Critical Events Analysis by Model (continued)}
\label{tab:17}
\small
\setlength{\tabcolsep}{5pt}
\renewcommand{\arraystretch}{1.3}

\begin{tabular}{|p{1.6cm}|p{14.4cm}|}
\hline
\multicolumn{2}{|c|}{\cellcolor{gray!20}\textbf{Molecular Machines}} \\
\hline
\multicolumn{2}{|l|}{\cellcolor{gray!10}\textbf{o3}} \\
\hline
\multicolumn{2}{|l|}{\textit{Model Label:} Foundation-First} \\
\hline
& \quad 1. 1960 s – Birth of supramolecular chemistry (Pedersen’s crown-ethers)  \\
& \quad 2.  1983 – Sauvage’s Cu(I)-templated catenane synthesis  \\
& \quad 3.  1991 – Stoddart’s donor–acceptor rotaxane \& molecular shuttle 
 \\
& \quad 4. 1999 – Feringa’s light-driven unidirectional molecular motor    \\
& \quad 5. 2016 – Award of the Nobel Prize for Molecular Machines  \\
\hline

\multicolumn{2}{|l|}{\cellcolor{gray!10}\textbf{Gemini 2.5 Pro}} \\
\hline
\multicolumn{2}{|l|}{\textit{Model Label:}  Enabling Function } \\
\hline
& \quad 1.Sauvage's Template-Directed Catenane Synthesis (1983)  \\
& \quad 2. Feringa's Light-Driven Unidirectional Molecular Motor (1999)  \\
& \quad 3. Stoddart's Controllable Molecular Shuttle (1991)  \\
& \quad 4. Sauvage Group's Demonstration of Relative Motion in Catenanes (post-1983)   \\
& \quad 5. The 2016 Nobel Prize Award  \\
\hline

\multicolumn{2}{|l|}{\cellcolor{gray!10}\textbf{Claude Sonnet 3.7 Thinking}} \\
\hline
\multicolumn{2}{|l|}{\textit{Model Label:} Gateway Thinking} \\
\hline
& \quad 1. Sauvage's Template-Directed Synthesis (1983)  \\
& \quad 2. Stoddart's Molecular Shuttle (1991)  \\
& \quad 3. Feringa's Unidirectional Molecular Motor (1999) \\
& \quad 4. Emergence of Supramolecular Chemistry (1960s-70s) \\
& \quad 5. Practical Macroscopic Demonstrations (Early 2000s) \\
\hline
\end{tabular}
\end{table*}

\begin{table*}[htbp]
\centering
\caption{Scientific Discovery Critical Events Analysis by Model (continued)}
\label{tab:18}
\small
\setlength{\tabcolsep}{5pt}
\renewcommand{\arraystretch}{1.3}

\begin{tabular}{|p{1.6cm}|p{14.4cm}|}
\hline
\multicolumn{2}{|c|}{\cellcolor{gray!20}\textbf{Generalization of Order Phenomena Methods}} \\
\hline
\multicolumn{2}{|l|}{\cellcolor{gray!10}\textbf{o3}} \\
\hline
\multicolumn{2}{|l|}{\textit{Model Label:} Analogy-Driven} \\
\hline
& \quad 1. 1972 – Polymer-Magnet n → 0 Analogy   \\
& \quad 2.  1971 – Landau-de Gennes Theory of the Nematic Transition  \\
& \quad 3.  1968 – Strategic Pivot to Liquid Crystals \& Orsay Group Launch  \\
& \quad 4. 1972 – Superconductor-Smectic A Analogy \& TGB Prediction   \\
& \quad 5.  1974/1979 – Twin Monographs Disseminate the Framework  \\
\hline

\multicolumn{2}{|l|}{\cellcolor{gray!10}\textbf{Gemini 2.5 Pro}} \\
\hline
\multicolumn{2}{|l|}{\textit{Model Label:} Analogical Insight } \\
\hline
& \quad 1. Sequential Immersion in Magnetism and Superconductivity  \\
& \quad 2. Shift of Focus to Liquid Crystals (~1968)  \\
& \quad 3. Polymer-Magnet Analogy (n=0 Theorem)  \\
& \quad 4. Landau-de Gennes Theory for Nematic-Isotropic Transition  \\
& \quad 5. Superconductor-Smectic A Analogy   \\
\hline

\multicolumn{2}{|l|}{\cellcolor{gray!10}\textbf{Claude Sonnet 3.7 Thinking}} \\
\hline
\multicolumn{2}{|l|}{\textit{Model Label:} Boundary Transcendence} \\
\hline
& \quad 1. De Gennes' sequential immersion in magnetism, superconductivity, and finally complex materials (1955-1971)  \\
& \quad 2. The Superconductor-Smectic A Analogy (1972) \\
& \quad 3. The Polymer-Magnet Analogy/n=0 Theorem (1972) \\
& \quad 4. Establishment of the Orsay Liquid Crystal Group (~1968) \\
& \quad 5. Publication of landmark books on liquid crystals (1974) and polymer physics (1979)  \\
\hline
\end{tabular}
\end{table*}

\begin{table*}[htbp]
\centering
\caption{Scientific Discovery Critical Events Analysis by Model (continued)}
\label{tab:19}
\small
\setlength{\tabcolsep}{5pt}
\renewcommand{\arraystretch}{1.3}

\begin{tabular}{|p{1.6cm}|p{14.4cm}|}
\hline
\multicolumn{2}{|c|}{\cellcolor{gray!20}\textbf{Genomes of Extinct Hominins and Human Evolution}} \\
\hline
\multicolumn{2}{|l|}{\cellcolor{gray!10}\textbf{o3}} \\
\hline
\multicolumn{2}{|l|}{\textit{Model Label:} Barrier-Breaking} \\
\hline
& \quad 1.  Clean-room \& authentication protocols established (early 1990s)  \\
& \quad 2.  Ancient-DNA-tailored next-generation sequencing adopted (2005-2008)  \\
& \quad 3. Draft Neanderthal nuclear genome published (Science, 2010)   \\
& \quad 4.  Genetic discovery of Denisovans from a finger bone (Nature, 2010)  \\
& \quad 5. First Neanderthal mtDNA sequence authenticated (Cell, 1997)   \\
\hline

\multicolumn{2}{|l|}{\cellcolor{gray!10}\textbf{Gemini 2.5 Pro}} \\
\hline
\multicolumn{2}{|l|}{\textit{Model Label:} Enabling Breakthroughs} \\
\hline
& \quad 1. Adoption and Adaptation of Next-Generation Sequencing (NGS)  \\
& \quad 2. Publication of the Draft Neanderthal Nuclear Genome (2010)  \\
& \quad 3. Systematic Development of Stringent Anti-Contamination Methods  \\
& \quad 4. Discovery of the Denisovans from Genetic Data Alone (2010)  \\
& \quad 5. Sequencing of the First Neanderthal Mitochondrial DNA (mtDNA) (1997)  \\
\hline

\multicolumn{2}{|l|}{\cellcolor{gray!10}\textbf{Claude Sonnet 3.7 Thinking}} \\
\hline
\multicolumn{2}{|l|}{\textit{Model Label:} Methodical Revolutionary} \\
\hline
& \quad 1. Development of rigorous ancient DNA laboratory protocols (late 1990s-2000s)  \\
& \quad 2. Adoption of Next-Generation Sequencing technologies (mid-2000s)  \\
& \quad 3. Draft Neanderthal nuclear genome publication (2010)  \\
& \quad 4. Discovery of Denisovans (2010)
 \\
& \quad 5. Pääbo's pioneering work on Egyptian mummies (1985)
 \\
\hline
\end{tabular}
\end{table*}

\begin{table*}[htbp]
\centering
\caption{Scientific Discovery Critical Events Analysis by Model}
\label{tab:20}
\small
\setlength{\tabcolsep}{5pt}
\renewcommand{\arraystretch}{1.3}

\begin{tabular}{|p{1.6cm}|p{14.4cm}|}
\hline
\multicolumn{2}{|c|}{\cellcolor{gray!20}\textbf{Understanding Complex Systems}} \\
\hline
\multicolumn{2}{|l|}{\cellcolor{gray!10}\textbf{o3}} \\
\hline
\multicolumn{2}{|l|}{\textit{Model Label:} Keystone-Hunter} \\
\hline
& \quad 1. Manabe-Wetherald 1967 $CO_2$ + Water-Vapor Model   \\
& \quad 2.   Parisi 1979 Replica-Symmetry-Breaking Solution  \\
& \quad 3.  Hasselmann 1976 Stochastic Climate Framework  \\
& \quad 4.   Hasselmann 1993 Optimal Fingerprinting  \\
& \quad 5.  Manabe-Wetherald 1975 First 3-D GCM $CO_2$ Experiment  \\
\hline

\multicolumn{2}{|l|}{\cellcolor{gray!10}\textbf{Gemini 2.5 Pro}} \\
\hline
\multicolumn{2}{|l|}{\textit{Model Label:} Framework-Founding} \\
\hline
& \quad 1. Parisi's proposal of infinite Replica Symmetry Breaking (RSB) (1979)  \\
& \quad 2. Manabe's incorporation of key physical feedbacks in 1D models (1967)   \\
& \quad 3. Hasselmann's development of optimal fingerprinting methods (formalized 1993, 1997)   \\
& \quad 4.  Hasselmann's proposal of stochastic climate models (1976) \\
& \quad 5. The demonstrated failure of the Replica Symmetric (RS) solution for spin glasses (pre-1979)  \\
\hline

\multicolumn{2}{|l|}{\cellcolor{gray!10}\textbf{Claude Sonnet 3.7 Thinking}} \\
\hline
\multicolumn{2}{|l|}{\textit{Model Label:} Conceptual Integration} \\
\hline
& \quad 1. Manabe's 1967 Integration of Water Vapor Feedback into Climate Models \\
& \quad 2. Hasselmann's 1976 Introduction of Stochastic Climate Models \\
& \quad 3. Parisi's 1979 Discovery of Replica Symmetry Breaking \\
& \quad 4. Hasselmann's Development of "Fingerprinting" Methods (1979-1997) \\
& \quad 5. Mathematical Validation of Parisi's RSB Solution (1990s-2000s) \\
\hline
\end{tabular}
\end{table*}

\begin{table*}[htbp]
\centering
\caption{Scientific Discovery Critical Events Analysis by Model}
\label{tab:21}
\small
\setlength{\tabcolsep}{5pt}
\renewcommand{\arraystretch}{1.3}

\begin{tabular}{|p{1.6cm}|p{14.4cm}|}
\hline
\multicolumn{2}{|c|}{\cellcolor{gray!20}\textbf{Discovery of the Blackbody Form and Anisotropy of the Cosmic Microwave Background Radiation}} \\
\hline
\multicolumn{2}{|l|}{\cellcolor{gray!10}\textbf{o3}} \\
\hline
\multicolumn{2}{|l|}{\textit{Model Label:} Causal-Linchpin Lens} \\
\hline
& \quad 1. 1992 – COBE DMR detects 30 µK anisotropies in the CMB  \\
& \quad 2. 1990 – FIRAS nails the CMB’s black-body spectrum at 2.725 K  \\
& \quad 3. 1986 – Post-Challenger redesign rescues COBE  \\
& \quad 4. 1976 – NASA approves COBE and forms the Science Working Group  \\
& \quad 5. 1965 – Penzias \& Wilson’s 3 K “hiss” meets Dicke’s Big-Bang interpretation  \\
\hline

\multicolumn{2}{|l|}{\cellcolor{gray!10}\textbf{Gemini 2.5 Pro}} \\
\hline
\multicolumn{2}{|l|}{\textit{Model Label:} Critical Junctures} \\
\hline
& \quad 1. The Space Shuttle Challenger Disaster (1986) and Subsequent Redesign  \\
& \quad 2.  Penzias and Wilson's Accidental Discovery (1964)  \\
& \quad 3. Recognition of Ground/Balloon Experiment Limitations  \\
& \quad 4. DMR's Detection of CMB Anisotropies (Announced 1992)  \\
& \quad 5. FIRAS's Measurement of the Blackbody Spectrum (Announced 1990)  \\
\hline

\multicolumn{2}{|l|}{\cellcolor{gray!10}\textbf{Claude Sonnet 3.7 Thinking}} \\
\hline
\multicolumn{2}{|l|}{\textit{Model Label:} Adaptive Perseverance} \\
\hline
& \quad 1. Theoretical Prediction of CMB (1948) \\
& \quad 2. Discovery of CMB by Penzias and Wilson (1965) \\
& \quad 3. Space Shuttle Challenger Disaster (1986) \\
& \quad 4. FIRAS Spectrum Measurement (1990) \\
& \quad 5. DMR Anisotropy Detection (1992) \\
\hline
\end{tabular}
\end{table*}

\begin{table*}[htbp]
\centering
\caption{Scientific Discovery Critical Events Analysis by Model}
\label{tab:22}
\small
\setlength{\tabcolsep}{5pt}
\renewcommand{\arraystretch}{1.3}

\begin{tabular}{|p{1.6cm}|p{14.4cm}|}
\hline
\multicolumn{2}{|c|}{\cellcolor{gray!20}\textbf{CRISPR Gene Editing}} \\
\hline
\multicolumn{2}{|l|}{\cellcolor{gray!10}\textbf{o3}} \\
\hline
\multicolumn{2}{|l|}{\textit{Model Label:} Mechanistic Leaps } \\
\hline
& \quad 1.  2012 – Jinek et al., Science \\
& \quad 2.Spring 2011 – Charpentier \& Doudna strike a collaboration in San Juan \\
& \quad 3. March 2011 – Charpentier lab uncovers tracrRNA–Cas9 mechanism  \\
& \quad 4. 2007 – Barrangou et al. demonstrate CRISPR adaptive immunity in bacteria  \\
& \quad 5. 1987 – Ishino et al. stumble on clustered repeats in E. coli  \\
\hline

\multicolumn{2}{|l|}{\cellcolor{gray!10}\textbf{Gemini 2.5 Pro}} \\
\hline
\multicolumn{2}{|l|}{\textit{Model Label:} Engineering Enablement} \\
\hline
& \quad 1. Engineering the sgRNA \& Demonstrating Programmability (Jinek et al., 2012)  \\
& \quad 2. Identification of tracrRNA and the S. pyogenes Mechanism (Charpentier et al., 2011)  \\
& \quad 3. Demonstration of Eukaryotic Cell Editing (Zhang/Church et al., 2013)  \\
& \quad 4. Experimental Proof of Adaptive Immunity (Barrangou et al., 2007)  \\
& \quad 5. The Charpentier-Doudna Collaboration (2011)  \\
\hline

\multicolumn{2}{|l|}{\cellcolor{gray!10}\textbf{Claude Sonnet 3.7 Thinking}} \\
\hline
\multicolumn{2}{|l|}{\textit{Model Label:} Discovery to Tool} \\
\hline
& \quad 1. Jinek et al. Science Paper (2012) \\
& \quad 2. Discovery of tracrRNA (2011) \\
& \quad 3. Experimental Proof of CRISPR's Immune Function (2007)\\
& \quad 4. Charpentier-Doudna Collaboration (2011) \\
& \quad 5. Demonstration in Eukaryotic Cells (2013) \\
\hline
\end{tabular}
\end{table*}

\begin{table*}[htbp]
\centering
\caption{Scientific Discovery Critical Events Analysis by Model}
\label{tab:23}
\small
\setlength{\tabcolsep}{5pt}
\renewcommand{\arraystretch}{1.3}

\begin{tabular}{|p{1.6cm}|p{14.4cm}|}
\hline
\multicolumn{2}{|c|}{\cellcolor{gray!20}\textbf{Cellular Oxygen Sensing}} \\
\multicolumn{2}{|l|}{\cellcolor{gray!10}\textbf{o3}} \\
\hline
\multicolumn{2}{|l|}{\textit{Model Label:} Mechanistic-Keystone} \\
\hline
& \quad 1. Prolyl-hydroxylation switch revealed (2001) \\
& \quad 2. VHL targets HIF-$\alpha$ for destruction (1999) \\
& \quad 3. HIF-1 purified and cloned as an $\rm O_2$-labile heterodimer (1995) \\
& \quad 4. Hypoxia Response Element mapped in the EPO gene (1991) \\
& \quad 5. $\rm O_2$-sensing machinery shown to be ubiquitous (1993) \\
\hline

\multicolumn{2}{|l|}{\cellcolor{gray!10}\textbf{Gemini 2.5 Pro}} \\
\hline
\multicolumn{2}{|l|}{\textit{Model Label:} Mechanism Matters Most} \\
\hline
& \quad 1.  Prolyl Hydroxylation Identified as the Oxygen-Dependent Switch (2001) \\
& \quad 2. VHL Linked to Oxygen-Dependent HIF-1$\alpha$ Degradation (1999)  \\
& \quad 3. Discovery of Widespread Oxygen Sensing Machinery (1993)  \\
& \quad 4. Biochemical Purification and Cloning of HIF-1$\alpha$ (1995) \\
& \quad 5. Identification of the Hypoxia Response Element (HRE) (1991-1992)  \\
\hline

\hline
\multicolumn{2}{|l|}{\cellcolor{gray!10}\textbf{Claude Sonnet 3.7 Thinking}} \\
\hline
\multicolumn{2}{|l|}{\textit{Model Label:} Mechanistic Convergence} \\
\hline
& \quad 1. Discovery of prolyl hydroxylation as the oxygen-sensing switch (2001)  \\
& \quad 2. Discovery that VHL targets HIF-$\alpha$ for oxygen-dependent degradation (1999)  \\
& \quad 3. Identification and purification of HIF-1 as a heterodimer (1992-1995)  \\
& \quad 4. Linking VHL disease to hypoxia-regulated genes (1996)  \\
& \quad 5.  Evidence for a widespread oxygen-sensing mechanism (1993)  \\
\hline
\end{tabular}
\end{table*}

\begin{table*}[!htbp]
\centering
\caption{Scientific Discovery Critical Events Analysis by Model}
\label{tab:24}
\small
\setlength{\tabcolsep}{5pt}
\renewcommand{\arraystretch}{1.3}

\begin{tabular}{|p{1.6cm}|p{14.4cm}|}
\hline
\multicolumn{2}{|c|}{\cellcolor{gray!20}\textbf{Gravitational wave detection}} \\
\hline
\multicolumn{2}{|l|}{\cellcolor{gray!10}\textbf{o3}} \\
\hline
\multicolumn{2}{|l|}{\textit{Model Label:} Critical Path} \\
\hline
& \quad 1. Einstein’s prediction of gravitational waves (1916)  \\
& \quad 2. Weiss’s laser-interferometer design \& noise analysis (1972)   \\
& \quad 3. NSF commitment and Barish-led LIGO reorganization (1990-1997)  \\
& \quad 4. Commissioning of Advanced LIGO (mid-2015)   \\
& \quad 5. GW150914 detection and 2016 announcement   \\
\hline

\multicolumn{2}{|l|}{\cellcolor{gray!10}\textbf{Gemini 2.5 Pro}} \\
\hline
\multicolumn{2}{|l|}{\textit{Model Label:} Enabling Breakthroughs} \\
\hline
& \quad 1. Weiss's 1972 Interferometer Proposal \& Noise Analysis  \\
& \quad 2. Advanced LIGO Upgrades (c. 2008-2015) \\
& \quad 3. LIGO Project Formation \& NSF Funding (1984-1991)  \\
& \quad 4. GW150914: First Direct Detection (2015)  \\
& \quad 5. Barish's Leadership \& LSC Formation (1994-1997)  \\
\hline

\multicolumn{2}{|l|}{\cellcolor{gray!10}\textbf{Claude Sonnet 3.7 Thinking}} \\
\hline
\multicolumn{2}{|l|}{\textit{Model Label:} Essential Enablers} \\
\hline
& \quad 1. Einstein's Prediction (1916) \\
& \quad 2. Weiss's Interferometer Concept (1972) \\
& \quad 3. Barish's Leadership (1994-1997) \\
& \quad 4. NSF/Congress Funding Approval (1990-1992) \\
& \quad 5. Advanced LIGO Upgrades (2008-2015) \\
\hline
\end{tabular}
\end{table*}

\begin{table*}[!htbp]
\centering
\caption{Scientific Discovery Critical Events Analysis by Model}
\label{tab:25}
\small
\setlength{\tabcolsep}{5pt}
\renewcommand{\arraystretch}{1.3}

\begin{tabular}{|p{1.6cm}|p{14.4cm}|}
\hline
\multicolumn{2}{|c|}{\cellcolor{gray!20}\textbf{X-ray Mutagenesis}} \\
\hline
\multicolumn{2}{|l|}{\cellcolor{gray!10}\textbf{o3}} \\
\hline
\multicolumn{2}{|l|}{\textit{Model Label:} Causal Pivots} \\
\hline
& \quad 1. Nov 1926–1927 X-ray mutation experiments succeed  \\
& \quad 2. Creation of the CIB lethal-screen stock (1926)   \\
& \quad 3. Strategic switch to controllable X-ray irradiation (late 1926) \\
& \quad 4. Full data reveal at the 5th International Genetics Congress, Berlin (Sept 1927)  \\
& \quad 5. Award of the 1946 Nobel Prize  \\
\hline

\multicolumn{2}{|l|}{\cellcolor{gray!10}\textbf{Gemini 2.5 Pro}} \\
\hline
\multicolumn{2}{|l|}{\textit{Model Label:} Enabling Proof} \\
\hline
& \quad 1. Development of the CIB Stock (c. 1926)  \\
& \quad 2. Execution of Definitive X-ray Experiments (Nov 1926 - early 1927)  \\
& \quad 3. Muller's Strategic Focus on Lethal Mutations for Quantification (pre-1926)  \\
& \quad 4. Presentation at the Berlin Congress (Sept 1927) \\
& \quad 5. Awarding of the Nobel Prize (1946)  \\
\hline

\multicolumn{2}{|l|}{\cellcolor{gray!10}\textbf{Claude Sonnet 3.7 Thinking}} \\
\hline
\multicolumn{2}{|l|}{\textit{Model Label:} Methodical Innovator} \\
\hline
& \quad 1. Definitive X-ray Experiments (November 1926)\\
& \quad 2. Development of the CIB Stock (1926) \\
& \quad 3. Development of Quantitative Methods for Detecting Lethal Mutations (1919) \\
& \quad 4.Theoretical Rationale for Testing X-rays (1926)\\
& \quad 5.  Presentation at the Fifth International Congress of Genetics (September 1927) \\
\hline
\end{tabular}
\end{table*}

\begin{table*}[!htbp]
\centering
\caption{Movie Script Critical Events Analysis by Model}
\label{tab:26}
\small
\setlength{\tabcolsep}{5pt}
\renewcommand{\arraystretch}{1.3}

\begin{tabular}{|p{1.6cm}|p{14.4cm}|}
\hline
\multicolumn{2}{|c|}{\cellcolor{gray!20}\textbf{\large Good Will Hunting}} \\
\hline

\multicolumn{2}{|l|}{\cellcolor{gray!10}\textbf{Phi-4}} \\
\hline
\multicolumn{2}{|l|}{\textit{Model Personality:} \textbf{Process-Oriented}} \\
\hline
& \quad 1. Therapy Sessions with Dr. Sean Maguire \\
& \quad 2. Lambeau's Recommendation to See a Therapist \\
& \quad 3. Will's Confrontation with Chuckie \\
& \quad 4. Lambeau's Job Offer from Tri-tech \\
& \quad 5. Will and Skylar's Relationship \\
\hline

\multicolumn{2}{|l|}{\cellcolor{gray!10}\textbf{Orca2}} \\
\hline
\multicolumn{2}{|l|}{\textit{Model Personality:} \textbf{Trauma-Focused}} \\
\hline
& \quad 1. Will confronts his foster father about the abuse he suffered, leading to a release of his pent-up anger and resentment. \\
& \quad 2. Skylar recognizes Will's talent for mathematics and encourages him to pursue his potential. \\
& \quad 3. Lambeau helps Will face his emotional trauma and overcome his self-destructive behavior. \\
& \quad 4. Will decides to leave his old life behind and commit to a relationship with Skylar, opening up new opportunities for himself. \\
& \quad 5. Will realizes that he has the ability to change his life and pursue higher education at MIT, fulfilling his potential.\\
\hline

\multicolumn{2}{|l|}{\cellcolor{gray!10}\textbf{Qwen 2.5}} \\
\hline
\multicolumn{2}{|l|}{\textit{Model Personality:} \textbf{Achievement-Oriented}} \\
\hline
& \quad 1. Will Solves Professor Lambeau's Challenging Mathematical Problem \\
& \quad 2. Professor Lambeau's Intervention \\
& \quad 3. Meeting Sean Maguire \\
& \quad 4. Will's Breakup with Skylar \\
& \quad 5. Will's Decision to Leave Boston for California \\
\hline
\end{tabular}
\end{table*}

\begin{table*}[!htbp]
\centering
\caption{Movie Script Critical Events Analysis by Model (continued)}
\label{tab:27}
\small
\setlength{\tabcolsep}{5pt}
\renewcommand{\arraystretch}{1.3}

\begin{tabular}{|p{1.6cm}|p{14.4cm}|}
\hline
\multicolumn{2}{|c|}{\cellcolor{gray!20}\textbf{\large Dark Knight Rises}} \\
\hline

\multicolumn{2}{|l|}{\cellcolor{gray!10}\textbf{Phi-4}} \\
\hline
\multicolumn{2}{|l|}{\textit{Model Personality:} \textbf{Conflict-Oriented}} \\
\hline
& \quad 1. Bane's Attack on Gotham City. \\
& \quad 2. Batman’s Imprisonment \\
& \quad 3. The Fall of the Bat Signal \\
& \quad 4. Selina Kyle (Catwoman) Rejoining Batman \\
& \quad 5. Final Battle in Gothtopia \\
\hline

\multicolumn{2}{|l|}{\cellcolor{gray!10}\textbf{Orca2}} \\
\hline
\multicolumn{2}{|l|}{\textit{Model Personality:} \textbf{Vulnerability-Focused}} \\
\hline
& \quad 1. Batman confronts Bane and his mercenaries on the street and gets severely beaten \\
& \quad 2. Alfred persuades Bruce to get out of his depression and live a normal life \\
& \quad 3. Bane breaks into the Blackgate prison and recruits the inmates to join his plan \\
& \quad 4. Miranda Tate reveals herself as Talia al Ghul's accomplice and activates the nuclear bomb \\
& \quad 5. Batman returns to the Batcave and escapes with Selina Kyle's help \\
\hline

\multicolumn{2}{|l|}{\cellcolor{gray!10}\textbf{Qwen 2.5}} \\
\hline
\multicolumn{2}{|l|}{\textit{Model Personality:} \textbf{Strategic}} \\
\hline
& \quad 1. Wayne's Capture by Bane \\
& \quad 2. Bane Reveals Bruce’s True Identity \\
& \quad 3. Daggett Seizes Control Over Wayne Enterprises \\
& \quad 4. The Battle Between Batman and Bane in Downtown Gotham \\
& \quad 5. The Final Stand Against Bane \\
\hline
\end{tabular}
\end{table*}

\begin{table*}[!htbp]
\centering
\caption{Movie Script Critical Events Analysis by Model (continued)}
\label{tab:28}
\small
\setlength{\tabcolsep}{5pt}
\renewcommand{\arraystretch}{1.3}

\begin{tabular}{|p{1.6cm}|p{14.4cm}|}
\hline
\multicolumn{2}{|c|}{\cellcolor{gray!20}\textbf{\large The Social Network}} \\
\hline

\multicolumn{2}{|l|}{\cellcolor{gray!10}\textbf{Phi-4}} \\
\hline
\multicolumn{2}{|l|}{\textit{Model Personality:} \textbf{Strategic}} \\
\hline
& \quad 1.Mark Zuckerberg's Decision to Create Facebook \\
& \quad 2. Involvement of Sean Parker as an Advisor \\
& \quad 3. Expansion of Facebook Beyond Ivy League Schools \\
& \quad 4. Confrontation with Eduardo Saverin and Accusations of Betrayal \\
& \quad 5. Legal Battle with the Winklevoss Twins \\
\hline

\multicolumn{2}{|l|}{\cellcolor{gray!10}\textbf{Orca2}} \\
\hline
\multicolumn{2}{|l|}{\textit{Model Personality:} \textbf{Origin-Focused}} \\
\hline
& \quad 1. Mark Zuckerberg creating Facemash \\
& \quad 2. Mark being sued by the Winklevoss twins. \\
& \quad 3. Mark's falling out with Eduardo Saverin \\
& \quad 4. Facebook's expansion to California and its impact on relationships \\
& \quad 5. The settlement between Mark and Eduardo \\
\hline

\multicolumn{2}{|l|}{\cellcolor{gray!10}\textbf{Qwen 2.5}} \\
\hline
\multicolumn{2}{|l|}{\textit{Model Personality:} \textbf{Talent-Recognition}} \\
\hline
& \quad 1. Creation of Facemash \\
& \quad 2. Facebook’s Launch \\
& \quad 3. Partnership with Eduardo Saverin \\
& \quad 4. Legal Battle with Winklevoss Twins \\
& \quad 5. Dilution of Shareholdings \\
\hline
\end{tabular}
\end{table*}

\begin{table*}[!htbp]
\centering
\caption{Movie Script Critical Events Analysis by Model (continued)}
\label{tab:29}
\small
\setlength{\tabcolsep}{5pt}
\renewcommand{\arraystretch}{1.3}

%
\end{table*}